\documentclass[10pt,twocolumn,letterpaper]{article}

\usepackage[pagenumbers]{iccv} %

\usepackage{graphicx}
\usepackage{booktabs}
\usepackage{dsfont}
\usepackage{multirow}
\usepackage{graphics}
\usepackage{wrapfig}
\usepackage{bbm}
\usepackage{xspace}
\usepackage{xcolor}
\usepackage{tikz}
\usepackage{pifont}

\newcommand{\methodName}{MambaMatcher\xspace}

\usepackage{xr}
\makeatletter
\newcommand*{\addFileDependency}[1]{%
  \typeout{(#1)}
  \@addtofilelist{#1}
  \IfFileExists{#1}{}{\typeout{No file #1.}}
}
\makeatother

\definecolor{iccvblue}{rgb}{0.21,0.49,0.74}
\usepackage[pagebackref,breaklinks,colorlinks,allcolors=iccvblue]{hyperref}

\def\paperID{2} %
\def\confName{ICCV}
\def\confYear{2025}

\title{Similarity-Aware Selective State-Space Modeling for Semantic Correspondence}

\author{Seungwook Kim \hspace{20mm} Minsu Cho \vspace{0.15cm}\\
Pohang University of Science and Technology (POSTECH), South Korea \\
{\tt\small \href{https://cvlab.postech.ac.kr/project/MambaMatcher/}{https://cvlab.postech.ac.kr/project/MambaMatcher/}}
}

\begin{document}

\maketitle

\begin{abstract}

Establishing semantic correspondences between images is a fundamental yet challenging task in computer vision. 
Traditional feature-metric methods enhance visual features but may miss complex inter-correlation relationships, while recent correlation-metric approaches are hindered by high computational costs due to processing 4D correlation maps.
We introduce \methodName, a novel method that overcomes these limitations by efficiently modeling high-dimensional correlations using selective state-space models (SSMs).
By implementing a similarity-aware selective scan mechanism adapted from Mamba's linear-complexity algorithm, \methodName refines the 4D correlation map effectively without compromising feature map resolution or receptive field. 
Experiments on standard semantic correspondence benchmarks demonstrate that \methodName achieves state-of-the-art performance.

\end{abstract}

\section{Introduction}
\label{sec:introduction}

Establishing semantic correspondences between images is a fundamental problem in computer vision, with wide-ranging applications in augmented and virtual reality (AR/VR)~\cite{kim2024enhancing}, virtual try-on~\cite{kim2023stableviton}, edit propagation~\cite{endo2016deepprop, peebles2022gan}, and instance swapping~\cite{zhang2024tale, park2024learning}. 
The task involves identifying semantically corresponding regions between pairs of images depicting different instances of the same class~\cite{cho2015unsupervised, min2019spair, truong2022probabilistic, min2023convolutional}.
Despite significant advancements in deep learning, reliably establishing semantic correspondences remains challenging due to substantial intra-class variations, including differences in pose, scale, and appearance.

Current state-of-the-art methods predominantly adopt a \textit{feature-metric approach}, enhancing the quality of dense features extracted from the images.
This enhancement is achieved by either (i) employing powerful feature extractors~\cite{tang2023emergent, zhang2024tale} or (ii) improving feature representations through additional convolutional or attention layers~\cite{seo2018attentive, lee2021learning, luo2024diffusion}.
While feature-metric methods can robustly identify semantic similarities across local pixels, they may struggle to capture complex inter-correlation relationships.

An alternative is the \textit{correlation-metric approach}, where the methods aim to model and refine inter-correlation relationships by processing the high-order correlation map constructed using the extracted image features~\cite{rocco2018neighbourhood, min2021convolutional, cho2021cats, kim2022transformatcher}.
Although this approach can alleviate ambiguous or noisy correspondences by considering global correlation patterns, processing the 4D correlation map incurs up to quartic complexity with respect to the feature map dimensions.
This severely limits the feature map resolution and necessitates compromises on the receptive field or network expressivity - which are critical factors for accurate and robust correspondences.
Consequently, despite their potential, correlation-metric methods are often outperformed by feature-metric methods that utilize stronger backbones and higher-resolution images~\cite{luo2024diffusion, hedlin2024unsupervised, li2023sd4match, tang2023emergent, zhang2024tale}.

In this paper, we propose \methodName, a novel approach that overcomes the limitations of both feature-metric and correlation-metric methods by efficiently modeling high-dimensional correlation maps using selective state-space models (SSMs).
To the best of our knowledge, \methodName is the first method to treat multi-level correlation scores at each position in the correlation map as a state in a state-space model, enabling effective and efficient modeling of inter-image correlations.
At the core of \methodName is a \textbf{\textit{similarity-aware selective scan}} mechanism, which adapts Mamba's linear selective scan algorithm to refine the 4D correlation map.
This mechanism allows us to robustly and scalably process the correlation map without compromising feature map resolution or receptive field, thereby capturing rich inter-image relationships while maintaining computational efficiency.

The key contributions of our work are as follows:

\begin{itemize} 
    \item We introduce \methodName, the first method to model high-order correlation maps using selective state-space models by treating multi-level correlation scores at each position as states. 
    
    \item We propose a novel similarity-aware selective scan mechanism, enabling efficient and accurate mining of inter-correlation relationships at high resolutions. 
    
    \item \methodName seamlessly integrates feature-metric and correlation-metric approaches into a unified pipeline, leveraging the strengths of both methods without compromising feature resolution or receptive field. 
    
    \item Extensive experiments demonstrate that \methodName achieves state-of-the-art performance on standard semantic correspondence benchmarks.
\end{itemize}

\section{Related Work}
\label{sec:related_work}

\noindent
\smallbreak
\textbf{Feature-metric semantic correspondence} prioritizes producing high-quality features to establish robust correspondences. Traditional methods~\citep{liu2010sift, bristow2015dense, cho2015unsupervised, ham2017proposal} typically use handcrafted descriptors~\citep{lowe2004distinctive, dalal2005histograms, bay2006surf} which are simple yet effective.
Recent methods demonstrate that using local features extracted from deep neural networks leads to significant performance improvements~\citep{min2020learning, tang2023emergent, luo2024diffusion, li2023sd4match}.
While ResNets~\citep{he2016deep} were the conventional choice for the visual feature extractor, more recent works propose employing stronger feature extractors such as DINOv2~\citep{oquab2023dinov2} or Stable Diffusion~\citep{rombach2022high}. 
In the presence of supervision, there are attempts to yield richer features by refining the extracted features, \textit{e.g.}, by using additional convolutional or attention layers~\citep{seo2018attentive, lee2021learning, huang2022learning}.
In our method, we leverage DINOv2 and refine its features using 2D convolutional layers tailored to enhance correspondence accuracy.
However, \methodName takes a step further by harmoniously integrating the correlation-metric approach through our proposed similarity-aware selective scan.

\noindent
\smallbreak
\textbf{Correlation-metric semantic correspondence} aims to refine ambiguities in the correlation map to establish more robust and accurate correspondences. 
NCNet~\citep{rocco2018neighbourhood} established this idea via a 4D convolutional network to consider neighborhood consensus, which motivated follow-up work to formulate more effective or efficient ways~\citep{li2020correspondence, lee2021patchmatch, kim2024hccnet}.
However, high-dimensional convolutional kernels are constrained by their local receptive field and static transformations.
To address this, CATs~\citep{cho2021cats, cho2022catspp} and TransforMatcher~\citep{kim2022transformatcher} apply self-attention to the correlation maps to consider inter-correlation relations in a dynamic global fashion.
However, applying self-attention to the correlation map incurs up to quartic computational complexity with respect to the feature map dimensions.
This necessitates a compromise on either the feature map dimensions, the receptive field of inter-correlation relationship mining, or the expressivity of the algorithm used, leading to sub-optimal results.
In our work, we introduce a novel approach that models the correlation space using selective state-space models, applying a similarity-aware selective scan to the correlation maps.
Building on the efficiency and scalability of Mamba~\citep{gu2023mamba}, this method effectively overcomes previous limitations by avoiding compromises on feature map dimensions, receptive field, or network expressivity. 

\noindent
\smallbreak
\textbf{State-space models for computer vision}
gained rapid interest with the advent of Mamba~\citep{gu2023mamba}, which showed promising results compared to transformer-based architectures for sequence modeling in natural language processing.
Notably, Mamba exhibits linear computational complexity at inference, in contrast to attention-based methods that typically have quadratic complexity.
VMamba~\citep{liu2024vmamba}, Vision Mamba~\citep{zhu2024vision}, and PlainMamba~\citep{yang2024plainmamba} concurrently propose adapting the selective scan algorithm to the 2D image domain by varying the scan directions to accommodate spatial dimensions.
These endeavors show competitive or superior performance compared to existing methods in the computer vision domain, motivating the application of Mamba to various downstream vision tasks~\citep{li2024videomamba, chen2024video,yue2024medmamba, ruan2024vm,liu2024point}. 
In our work, we extend Mamba's selective scan algorithm by introducing a similarity-aware selective scan specifically designed to refine 4D correlation maps.
This adaptation enables us to effectively model the correlation space using selective state-space models, allowing for seamless handling of high-dimensional data in semantic correspondence tasks. 

\begin{figure*}[t]
    \centering
    \includegraphics[width=1.0\textwidth]{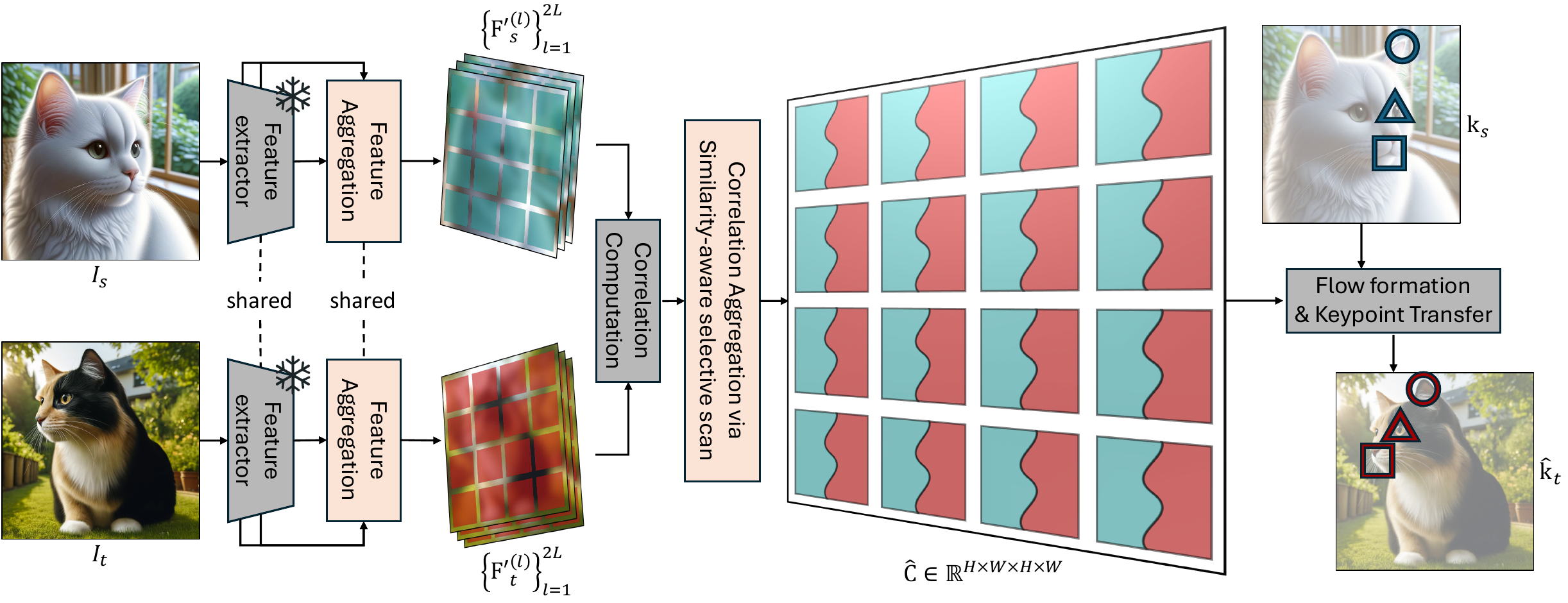}
    \vspace{-0.5cm}
    \caption{\textbf{Overview of \methodName}. 
    The multi-level correlation map is processed using our proposed similarity-aware selective scan mechanism to yield a refined correlation map $\hat{\mathbf{C}}$, which can be used to establish semantic correspondences between the images.
}
    \label{fig:method_overview}
    \vspace{-0.3cm}
\end{figure*}

\section{Preliminary: Selective State Space Models}
\label{sec:preliminary}

State-space models (SSM) can be viewed as linear time-invariant (LTI) systems that maps a 1D function or sequence $x(t) \in \mathbb{R} \mapsto y(t)$ through a hidden state $h(t) \in \mathbb{R}^\mathbf{N}$.
These models are mathematically formulated as linear ordinary differential equations (ODEs), with weighting parameters of $\mathbf{A} \in \mathbb{R}^{\mathbf{N} \times \mathbf{N}}$, $\mathbf{B} \in \mathbb{R}^{\mathbf{N} \times 1}, \mathbf{C} \in \mathbb{R}^{1 \times \mathbf{N}}$ and $D \in \mathbb{R}$:
\begin{equation}
\begin{split}
h'(t) &= \mathbf{A}h(t) + \mathbf{B}x(t), \\
y(t) &= \mathbf{C}h(t) + Dx(t)
\label{eqn:ssm_basic}
\end{split}
\end{equation}
Recently, the key idea is to use the HiPPO matrix~\citep{gu2020hippo} for $\mathbf{A}$, which produces a hidden state that memorizes the sequence history.
This is accomplished by tracking the coefficients of a Legendre polynomial, allowing the HiPPO matrix to approximate all of the previous history. 

The S4 and Mamba are based on discrete versions of Eq.\ref{eqn:ssm_basic}, which include a timescale parameter $\Delta$ to transform the continuous parameters $\mathbf{A}, \mathbf{B}$ to discrete parameters $\overline{\mathbf{A}}, \overline{\mathbf{B}}$.
The zero-order hold (ZOH) is commonly used for this transformation, where the discretized result is:
\begin{equation}
\begin{split}
    &\overline{\mathbf{A}} = \textrm{exp}(\Delta\mathbf{A}) \\
    &\overline{\mathbf{B}} = (\Delta\mathbf{A})^{-1}(\textrm{exp}(\Delta\mathbf{A}) - \mathbf{I}) \cdot \Delta\mathbf{B}
\end{split}
\label{eqn:zoh_discretization}
\end{equation}
Consequently, the discretized version of Eq.\ref{eqn:ssm_basic} using a step size of $\Delta$ can be rewritten as:
\begin{equation}
\begin{split}
h_t =\overline{\mathbf{A}}h_{t-1} + \overline{\mathbf{B}}x_t, y_t =\mathbf{\mathbf{C}}h_t + Dx_t
\end{split}
\end{equation}

S4~\citep{gu2021efficiently} uses input-independent matrices $\mathbf{A}$, $\mathbf{B}$, and $\mathbf{C}$, allowing parallel computation via convolutional reformulation. However, this input-independence limits S4's efficacy compared to dynamic, input-dependent self-attention mechanisms. To overcome this, Mamba~\citep{gu2023mamba} introduces input dependency by making $\mathbf{B}$, $\mathbf{C}$, and the step size $\Delta$ functions of the input, allowing the model to dynamically adapt and enhancing effectiveness over static models. This content-awareness, termed \textit{selective state-space models}, bridges the gap between the efficiency of state-space models and the adaptability of self-attention. Although this precludes convolution representations with fixed kernels, Mamba achieves parallelization via a parallel scan algorithm based on associative scan algorithms~\citep{martin2017parallelizing, smith2022simplified}. This leads to the \textit{selective scan algorithm}, which dynamically and efficiently scales linearly with sequence length, offering unbounded context and fast training/inference.

\begin{figure*}[t]
    \centering
    \includegraphics[width=1.0\textwidth]{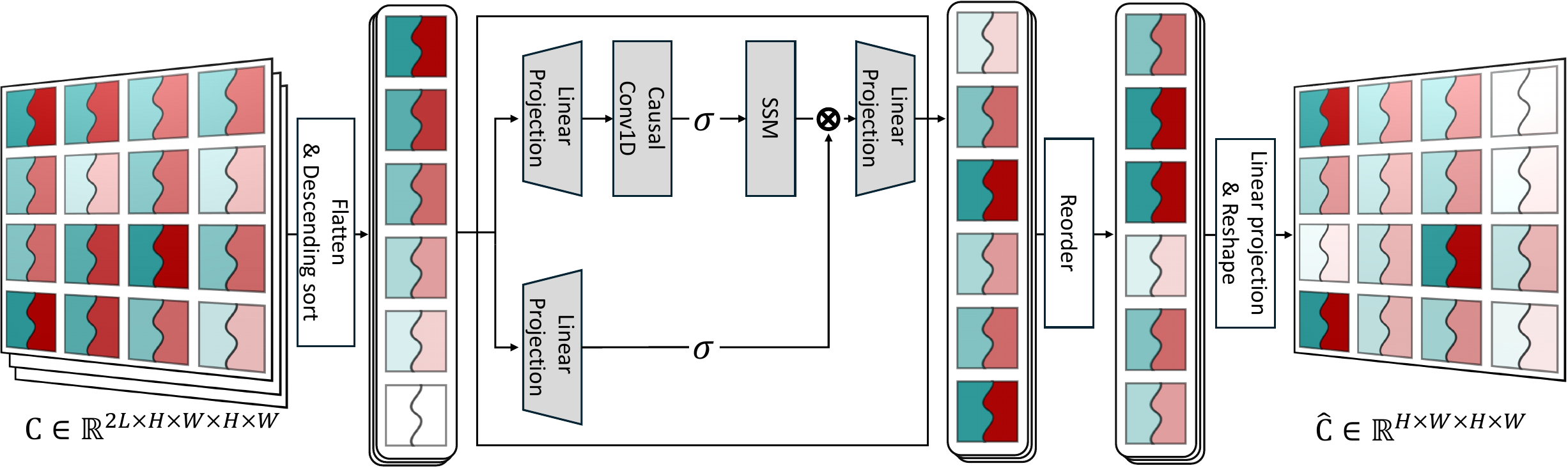}
    \caption{\textbf{Correlation aggregation via similiarity-aware selective scan}. 
    The multi-level correlation sequence is sorted in the descending order of similarity scores, such that the selective scan can be performed in a \textit{similarity-aware} manner.
    }
    \label{fig:mamba_details}
    \vspace{-2mm}
\end{figure*}

\section{MambaMatcher}
\label{sec:method}
We provide an overview of \methodName in Fig.~\ref{fig:method_overview}. Given a pair of images, we first extract multi-level feature maps from both images using a visual feature extractor. We then enhance these features using a simple yet effective convolutional feature aggregation module (Sec.~\ref{subsec:feature_extraction_aggregation}). Next, we construct a multi-level correlation map from these features. This correlation map is refined using our correlation aggregation layers based on our novel similarity-aware selective scan (Sec.~\ref{subsec:correlation_computation_aggregation}). Using the ground-truth source keypoints, we transfer them through the refined correlation map to obtain the \textit{predicted} target keypoints (Sec.~\ref{subsec:keypoint_transfer}). Finally, we train the entire network by comparing these predicted keypoints with the \textit{ground-truth} target keypoints (Sec.~\ref{subsec:training_objective}).

\subsection{Multi-level Feature Aggregation}
\label{subsec:feature_extraction_aggregation}

\noindent
\textbf{Multi-level Feature Extraction.}
Given a pair of images $(I_s, I_t)$, we use the pretrained DINOv2 ViT-B/14~\citep{oquab2023dinov2} to extract multi-level features from both the token and value representations across $L$ intermediate layers, yielding $2L$ sets of features for each of the source and target images, i.e., $\{(\mathbf{F}^{(l)}_s, \mathbf{F}^{(l)}_t)\}_{l=1}^{2L}$.

\smallbreak
\noindent
\textbf{Multi-level Feature Aggregation.}
We enhance the multi-level features through our 2D convolution-based feature aggregation layers to improve their self-awareness and robustness.
Formally, for each level $l$, the feature aggregation process is defined as:
\begin{align}
\mathbf{F}'^{(l)} &= \sigma\big(\mathbf{W}_2 * \big(\sigma(\mathbf{W}_1 * \mathbf{F}^{(l)})\big)\big)
\end{align}
where $\mathbf{W}_1$ and $\mathbf{W}_2$ are convolutional kernels,
and $\sigma(\cdot)$ is an activation function.
As a result, we obtain $2L$ sets of aggregated features, $\{(\mathbf{F}'^{(l)}_s, \mathbf{F}'^{(l)}_t)\}_{l=1}^{2L}$. 
The feature aggregators share the same weights across all levels.

\subsection{Multi-level Correlation Aggregation}
\label{subsec:correlation_computation_aggregation}

\noindent
\textbf{Multi-level Correlation Map Computation.}
Using the refined features from the previous stage, we compute a correlation map $\mathbf{C}^{(l)} \in \mathbb{R}^{H \times W \times H \times W}$ for each level $l$:
\begin{equation}
\mathbf{C}^{(l)}(p_s,p_t) = \frac{\mathbf{F}'^{(l)}_s(p_s) \cdot \mathbf{F}'^{(l)}_t(p_t)}{\| \mathbf{F}'^{(l)}_s(p_s) \| \| \mathbf{F}'^{(l)}_t(p_t) \|}
\end{equation}
where $p_s$ and $p_t$ denote spatial positions in the source and target feature maps, respectively, and $\|\cdot\|$ represents the L2 norm. The resulting $2L$ correlation maps are stacked to form a multi-level correlation map $\mathbf{C} \in \mathbb{R}^{2L \times H \times W \times H \times W}$.

\smallbreak
\noindent
\textbf{Correlation Aggregation via Similarity-aware Selective Scan.}
We flatten the multi-level correlation map $\mathbf{C}$ to form a \textit{correlation sequence} $\overline{\mathbf{C}} \in \mathbb{R}^{(H \times W \times H \times W) \times 2L}$, as selective state-space models are adept at handling sequential data. Here, the $2L$ channels correspond to the similarity scores from each level; we treat these multi-level similarity scores as the `state' in the selective SSM.

To embed similarity-awareness into the selective scan mechanism, we propose scanning the correlation sequence in descending order of similarity scores. 
This approach allows the model to initially establish a strong and reliable contextual foundation based on accurate matches, reducing interference from less certain correspondences. 
This is crucial for (1) \textbf{disambiguating high-similarity regions}, as early processing of strong matches helps resolve ambiguities, and (2) \textbf{refining low-similarity regions}, as ambiguous or noisy correspondences in later stages can leverage the context from earlier, more confident matches.

We sort the correlation sequence based on the similarity scores from the final correlation map (the $2L$-th level). After processing the sorted sequence with our similarity-aware selective scan mechanism, we reorder the sequence back to its original order. We then apply a linear projection and reshape to yield the refined correlation map $\hat{\mathbf{C}} \in \mathbb{R}^{H \times W \times H \times W}$.
Figure~\ref{fig:mamba_details} visualizes our correlation aggregation via similarity-aware selective scan.

\subsection{Keypoint Transfer}
\label{subsec:keypoint_transfer}

To transfer keypoints from the source image to the target image, we transform the refined correlation tensor $\hat{\mathbf{C}}$ into a dense flow field using the kernel soft-argmax technique~\citep{lee2019sfnet}. Specifically, for each source keypoint position $(i,j)$, we apply a 2D Gaussian kernel $\mathbf{G}^{\mathbf{p}}_{kl}$ centered at $\mathbf{p} = \arg\max_{k,l} \hat{\mathbf{C}}_{ijkl}$ to promote a unimodal matching probability distribution, mitigating erroneous transfers due to ambiguous matches.
We normalize the raw correlation outputs as follows:
\begin{equation}
\mathbf{C}^{\textrm{norm}}(i,j,k,l) = \frac{\exp\big(\mathbf{G}^{\mathbf{p}}_{kl} \hat{\mathbf{C}}(i,j,k,l)\big)}{\sum_{(k', l')} \exp\big(\mathbf{G}^{\mathbf{p}}_{k'l'} \hat{\mathbf{C}}(i,j,k',l')\big)}
\end{equation}
Using $\mathbf{C}^{\textrm{norm}}$, we transfer all coordinates on a dense grid $\mathbf{P} \in \mathbb{R}^{H \times W \times 2}$ to obtain their transferred coordinates $\hat{\mathbf{P}}$ on the target image $I_t$:
\begin{equation}
\hat{\mathbf{P}}(i,j) = \sum_{k,l} \mathbf{C}^{\textrm{norm}}(i,j,k,l) \cdot (k,l)
\end{equation}
Here, $(k,l)$ are spatial coordinates in the target image. $\mathbf{P}$ and $\hat{\mathbf{P}}$ are used to construct a dense flow field to transfer source keypoints $\mathbf{k}_s$ to predicted target keypoints $\hat{\mathbf{k}}_t$.

\subsection{Training Objective}
\label{subsec:training_objective}

Given an image pair $(I_s, I_t)$ with $M$ ground-truth keypoints, we use the above keypoint transfer scheme to obtain predicted target keypoints $\{\hat{\mathbf{k}}_t^{(m)}\}_{m=1}^M$. Our training objective is to minimize the average Euclidean distance between the predicted and ground-truth target keypoints:
\begin{equation}
\mathcal{L}_{\textrm{kp}} = \frac{1}{M} \sum_{m=1}^M \| \hat{\mathbf{k}}_t^{(m)} - \mathbf{k}_t^{(m)} \|^2_2
\end{equation}
Despite the simplicity of this loss function, our method achieves superior performance due to the effectiveness of the refined correlation map.

\begin{table*}
\vspace{-2.5mm}
  \caption{\textbf{Evaluation on PF-PASCAL and SPair-71k datasets.} \methodName outperforms existing baselines on both datasets, with reasonable latency and memory usage.
  We detail the backbone, supervision, and data augmentation of each method in Appendix~\ref{sec:appendix_baseline_details}.}
  \label{tbl:pfpascal_spair}
  \centering
  \resizebox{0.90\textwidth}{!}{%
  \begin{tabular}{llcccccccc}
    \toprule
    \multirow{3}{*}{Method} & \multirow{3}{*}{Image res.} & \multicolumn{3}{c}{PF-PASCAL} & \multicolumn{3}{c}{SPair-71k} & \multirow{3}{*}{\shortstack{time\\(\emph{ms})}} & \multirow{3}{*}{\shortstack{memory\\(GB)}} \\

    & & \multicolumn{3}{c}{@$\alpha_{\text{img}}$} & \multicolumn{3}{c}{@$\alpha_{\text{bbox}}$} \\
    \cmidrule(r){3-5}
    \cmidrule(r){6-8}
    & & 0.05 & 0.10  & 0.15 & 0.05 & 0.10 & 0.15 \\
    \midrule
    PWarpC-NCNet~\citeyearpar{truong2022probabilistic} & 400$\times$400 & 79.2 & 92.1 & 95.6 & 31.6 & 52.0 & 61.8 & - & - \\
    TransforMatcher~\citeyearpar{kim2022transformatcher} & 240$\times$240 & 80.8 & 91.8 & - & 32.4 & 53.7 & - & 54 & 1.6 \\
    NeMF~\citeyearpar{hong2022neural} & 512$\times$512 & 80.6 & 93.6 & - & 34.2 & 53.6 & - & 8500 & 6.3 \\
    SCorrSAN~\citeyearpar{huang2022learning} & 256$\times$256 & 81.5 & 93.3 & - & - & 55.3 & - & 28 & 1.5  \\
    HCCNet~\citeyearpar{kim2024hccnet} & 240$\times$240 & 80.2 & 92.4 & - & 35.8 & 54.8 & - & 30 & 2.0 \\
    CATs++~\citeyearpar{cho2022catspp} & 512$\times$512 & 84.9 & 93.8 & 96.8 & 40.7 & 59.8 & 68.5 & - & - \\
    UFC~\citeyearpar{hong2023unifying} & 512$\times$512 & \textbf{88.0} & 94.8 & 97.9 & 48.5 & 64.4 & \underline{72.1} & - & - \\
    DIFT~\citeyearpar{tang2023emergent} & 768$\times$768 & 69.4 & 84.6 & 88.1 & 39.7 & 52.9 & - & - & - \\
    DINO+SD$_\textrm{zero-shot}$~\citeyearpar{zhang2024tale} & 840$^2$ / 512$^2$ & 73.0 & 86.1 & 91.1 & - & 64.0 & - & - & - \\
    DINO+SD$_\textrm{sup}$~\citeyearpar{zhang2024tale} & 840$^2$ / 512$^2$ & 80.9 & 93.6 & 96.9 & - & 74.6 & - & - & -  \\
    Diffusion Hyperfeatures~\citeyearpar{luo2024diffusion} & 224$\times$224 & - & 86.7 & - & - & 64.6 & - & 6620 & - \\
    ~\citet{hedlin2024unsupervised} & 0.93$\times$ori. & - & - & - & 28.9 & 45.4 & - & 90k$<$ & - \\
    SD4Match~\citeyearpar{li2023sd4match} & 768$\times$768 & 84.4 & \underline{95.2} & \underline{97.5} & \underline{59.5} & \underline{75.5} & - & - & -\\
    \midrule
    \methodName (Ours) & 420$\times$420 & \underline{87.3} & \textbf{95.9} & \textbf{98.2} & \textbf{61.6} & \textbf{77.8} & \textbf{84.3} & 74 & 2.1 \\
    \bottomrule
  \end{tabular}%
  }
\end{table*}

\begin{figure}[h]
    \centering
    \includegraphics[width=0.5\textwidth]{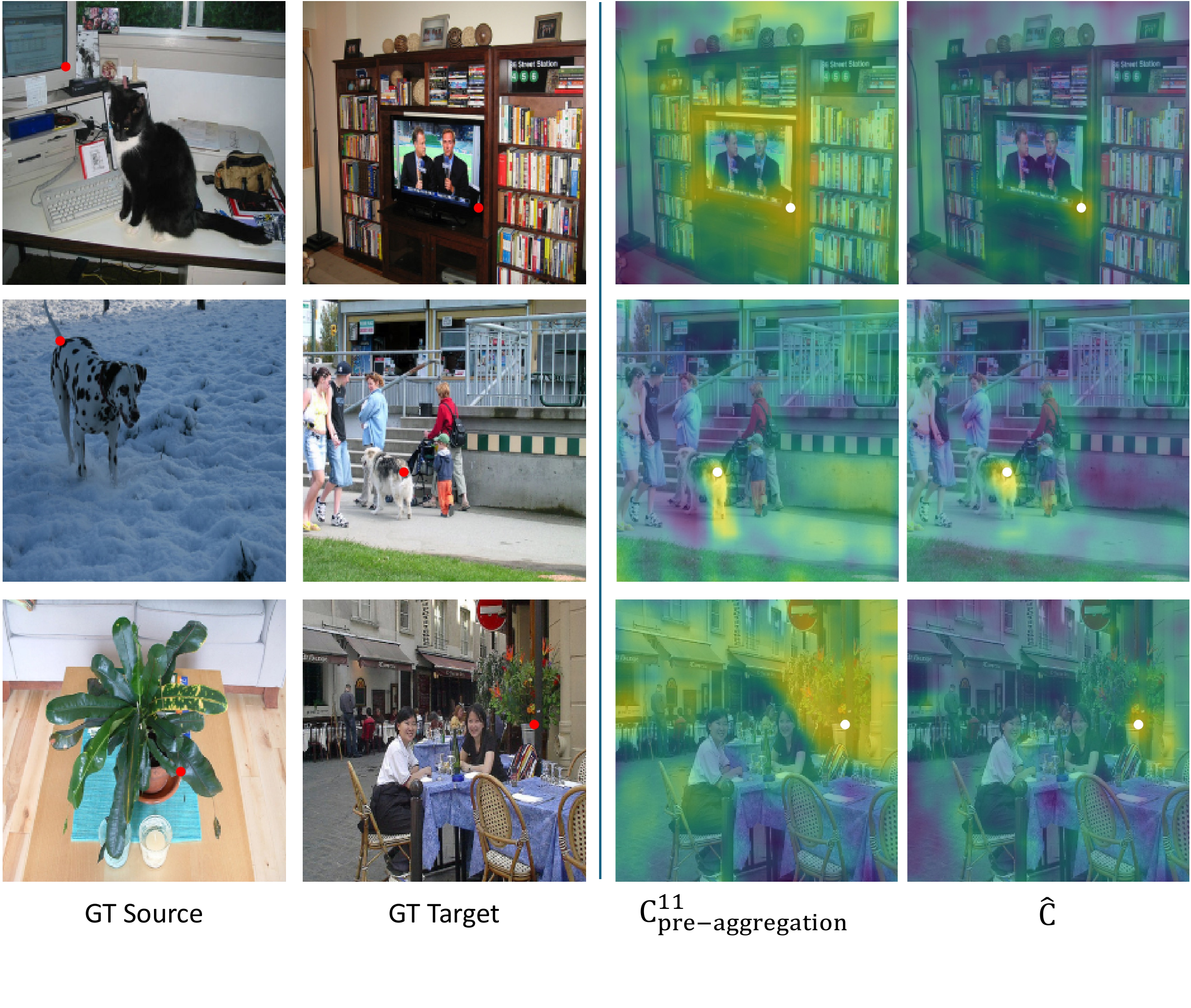}
    \vspace{-12mm}
    \caption{\textbf{Effect of similarity-aware selective scan.} 
    For each keypoint pair depicted in red, we visualize the initial and the final refined correlation.
    The refined correlation can better localize (\textit{i.e.,} has higher similarity, shown in brighter yellow) the keypoint.
    }
    \vspace{-5.0mm}
    \label{fig:mamba_viz}
\end{figure}

\begin{figure*}[ht]
    \centering
    \includegraphics[width=0.95\textwidth]{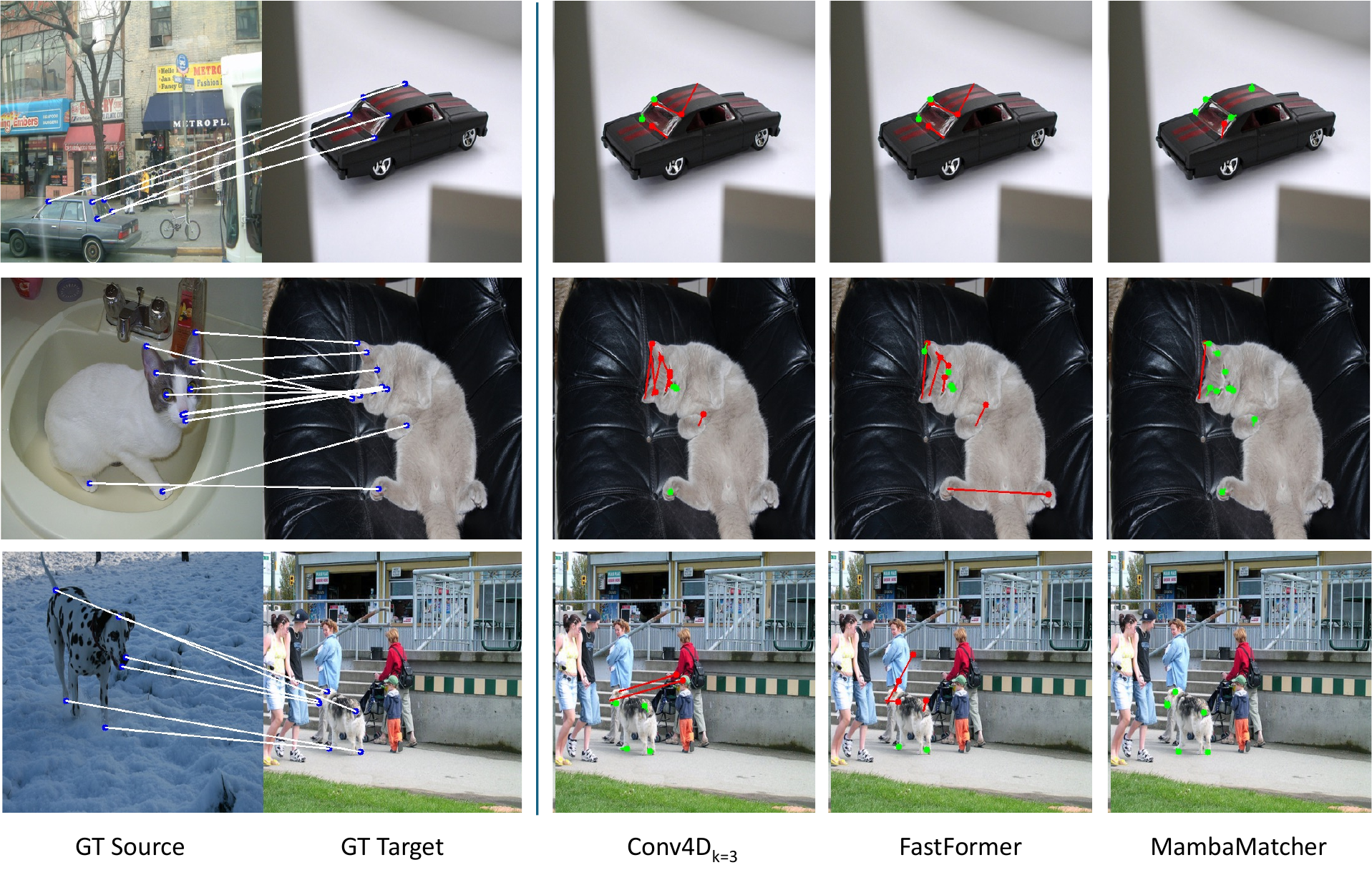}
    \caption{\textbf{Qualitative comparison to other correlation aggregation schemes}. Ground-truth correspondences are visualized in the left.
    The predicted keypoints are visualized on the right, where red depicts incorrect matches and green depicts correct matches.
    Our method shows to be more robust under large scale or viewpoint variations.
    Best viewed on electronics.}
    \label{fig:main_qual}
\end{figure*}

\section{Experiments}
\label{sec:experiments}
\noindent
\textbf{Implementation.}
We use DINOv2 (ViT-B/14)~\citep{oquab2023dinov2} to obtain local features. We resize input images to $420 \times 420$, resulting in feature maps of size $H = W = 30$ and correlation maps of size $30^4$. Considering that ViT-B/14 has 12 transformer layers, we extract the token and value representations from layers 4 to 11, yielding a total of $8$ layers $\times$ $2$ facets $=$ $16$ feature maps for each image. These feature maps serve as inputs to our subsequent feature aggregation layer.
Our feature aggregation layer consists of two layers of 2D convolution with a kernel size of $3$, having output channel dimensions of $4 \times 768 = 3072$ and $768$, respectively, with a ReLU activation function in between. For the correlation aggregation layer, we build upon the open-source implementation of Mamba~\citep{gu2023mamba}, using an SSM expansion factor of $16$, local convolution width of $4$, and block expansion factor of $3$.
We use the Adam optimizer~\citep{kingma2014adam} with a constant learning rate of $1\mathrm{e}{-3}$. We freeze the visual feature extractor during training to focus on learning the feature and correlation aggregation layers. 

\begin{table}[h]
\vspace{-5.0mm}
\centering
\begin{minipage}{0.48\textwidth}
\centering
    \vspace{+4mm}
\caption{\textbf{Single facet comparison.}  }
  \centering
  \begin{tabular}{cccc}
    \toprule
    \multirow{3}{*}{Facet} & \multicolumn{3}{c}{SPair-71k (s)} \\
    & \multicolumn{3}{c}{@$\alpha_{\text{bbox}}$}  \\
    \cmidrule(r){2-4}
    &  0.05 & 0.10 & 0.15\\

    \midrule
     Token & \textbf{25.2} & \textbf{43.1} & \textbf{55.7} \\
     Query & 18.6 & 33.3 & 43.3 \\
     Key & 15.6 & 30.0 & 41.4 \\
     Value & 24.4 & 41.8 & 53.6 \\
    \bottomrule
  \end{tabular}%
\label{tbl:feature_facet_ablation}

\end{minipage}%
\hfill%
\begin{minipage}{0.48\textwidth}
\centering
     \caption{\textbf{Facet combination comparison.}}
  \centering
  \begin{tabular}{cccc}
    \toprule
    \multirow{3}{*}{Facet used} & \multicolumn{3}{c}{SPair-71k (s)} \\
    & \multicolumn{3}{c}{@$\alpha_{\text{bbox}}$}  \\
    \cmidrule(r){2-4}
    &  0.05 & 0.10 & 0.15\\

    \midrule
    Token & 30.8 & 48.6 & 59.0 \\
    + Value & 32.3& 50.5 & 61.0 \\
    + Value, Query & 32.5 & 50.6 & 61.0 \\
    + Value, Query, Key & \textbf{33.2} & \textbf{50.7} & \textbf{61.3} \\
    \bottomrule
  \end{tabular}
  \label{tbl:feature_facet_multi_ablation}

\end{minipage}
\vspace{-3.0mm}
\end{table}

\begin{figure*}[h]
    \centering
    \includegraphics[width=0.95\textwidth]{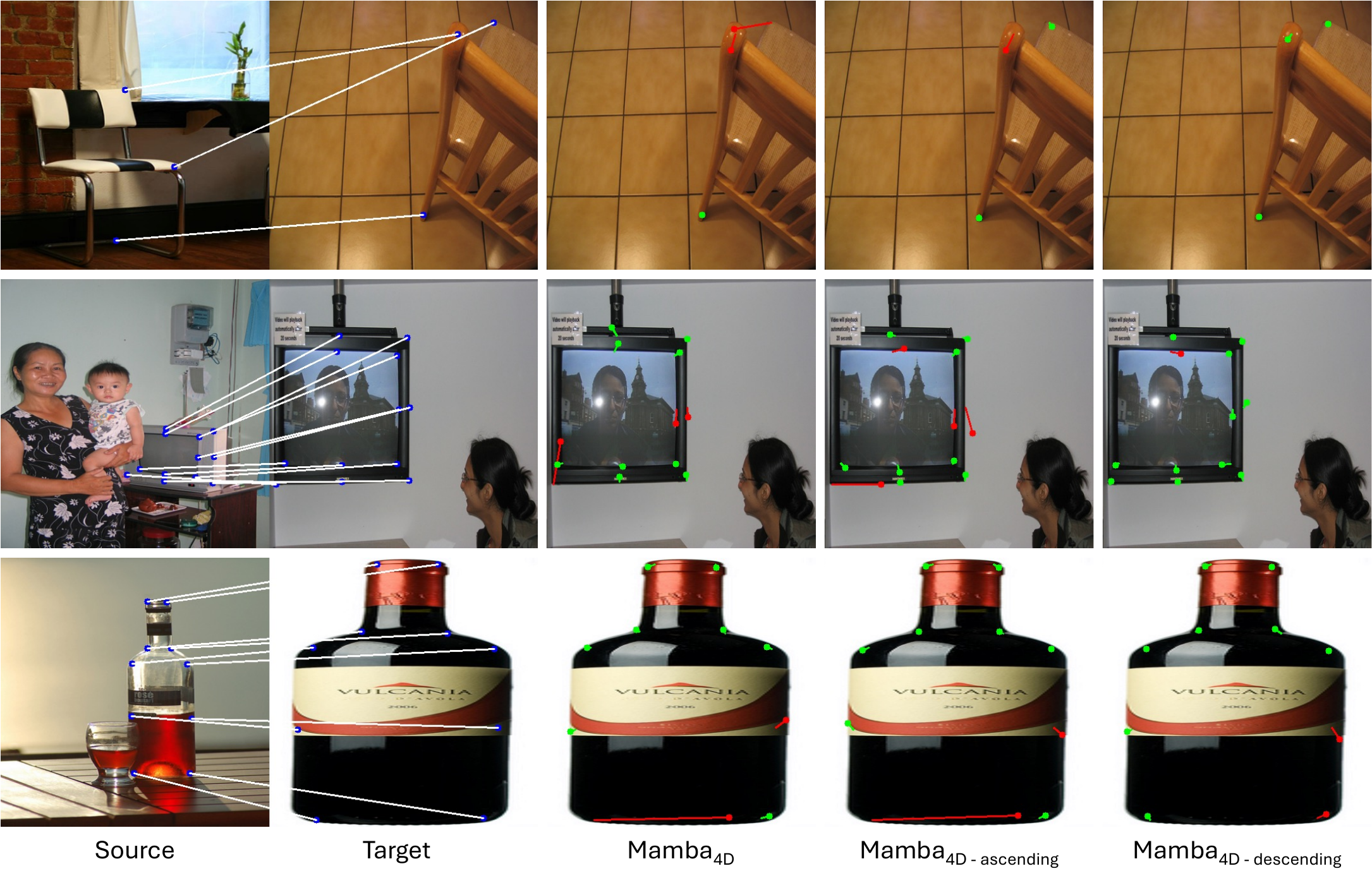} %
    \vspace{-3mm}
    \caption{\textbf{Comparison of Mamba$_{\textrm{4D}}$ scanning schemes.} It can be seen that our choice of scanning the correlation sequence in a descending order shows better keypoint localization, evidencing improved denoising and disambiguation of the correlation sequence.}
    \label{fig:scan_comparison}
\end{figure*}

\begin{figure*}
    \centering
    \includegraphics[width=\textwidth,height=\textheight,keepaspectratio]{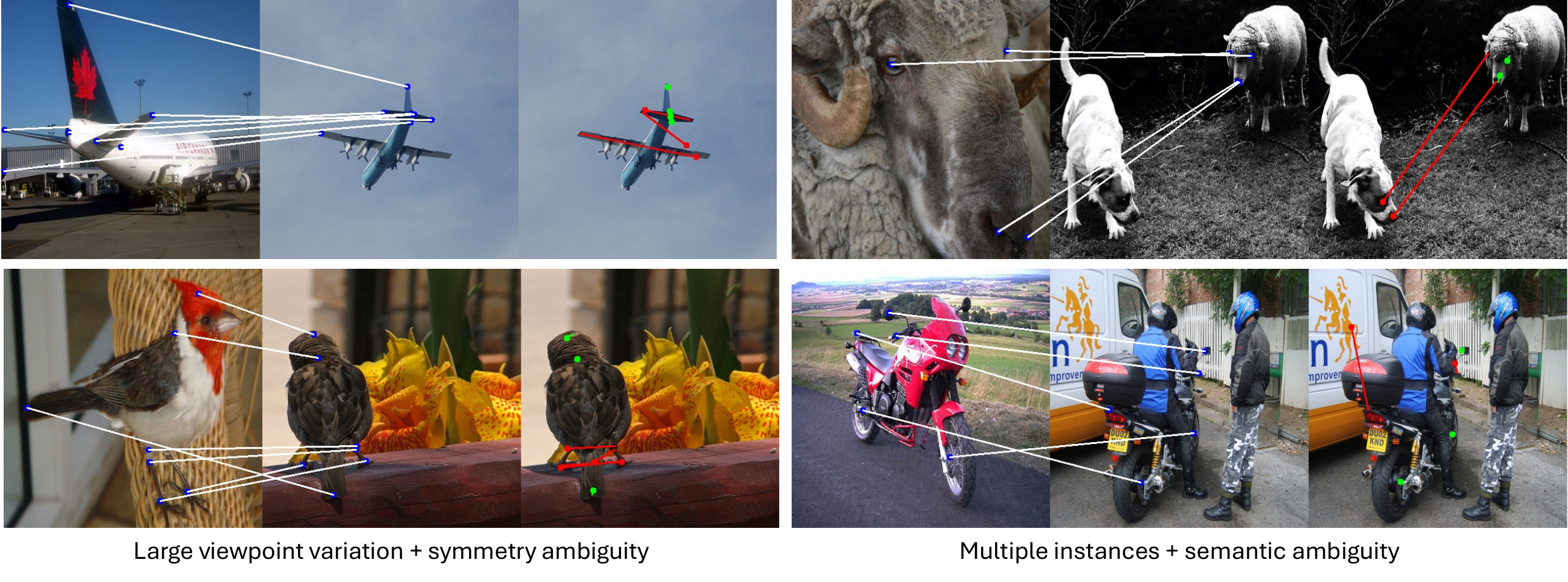} %
    \vspace{-5mm}
    \caption{\textbf{Failure case of MambaMatcher.} We analyze the common failure cases of our method. Firstly, MambaMatcher shows to fail more often in the dual presence of large viewpoint variation and symmetry ambiguity, where our model fails to accurately distinguish the position given a symmetric instance. Secondly, MambaMatcher often fails to follow the ground-truth in the dual presence of multiple instances and semantic ambiguity. For example, in the upper-right image, an eye and the nose of the sheep is predicted to correspond to an eye and the nose of a neighbouring dog.}
    \label{fig:failure_case}
\end{figure*}

\begin{table*}[h]
\vspace{-2.5mm}
\centering
\begin{minipage}{0.48\textwidth}
    \centering
      \caption{\textbf{Comparison between feature aggregation schemes.}
  }
  \centering
  \begin{tabular}{lccc}
    \toprule
    \multirow{3}{*}{Feature aggregation} & \multicolumn{3}{c}{SPair-71k (s)} \\
    & \multicolumn{3}{c}{@$\alpha_{\text{bbox}}$}  \\
    \cmidrule(r){2-4}
    & 0.05 & 0.10 & 0.15\\
    \midrule
    None & 32.3 & 50.5 & 61.0 \\
    \midrule
    2D Conv$_{k=1}$  & 54.5 & 72.3 & 79.8 \\
    2D Conv$_{k=3}$ & \underline{58.9} & \underline{77.9} & \underline{84.6} \\
    2D Conv$_{k=5}$& \textbf{59.2} & \textbf{78.4} & \textbf{85.3} \\
    \midrule
    Self-attn~\citeyearpar{dosovitskiy2020image} & 48.5 & 68.2 & 76.9 \\
    + Cross-attn. & 35.2 & 57.0 & 68.6 \\
    \midrule
    Mamba$_{\textrm{2D}}$~\citeyearpar{gu2023mamba} & 56.3 & 74.7 & 81.5 \\
    + bidirectional & 53.3 & 75.2 & 82.8 \\
    + Z-order curve~\citeyearpar{morton1966computer} & 56.0 & 75.2 & 82.8 \\
    PlainMamba~\citeyearpar{yang2024plainmamba} & 54.1 & 74.0 & 81.8 \\
    \bottomrule
  \end{tabular}%
\label{tbl:feature_aggregation_ablation}

\end{minipage}
\hfill
\begin{minipage}{0.48\textwidth}
    \centering
      \caption{\textbf{Comparison between correlation aggregation schemes.}}
  \centering
  \begin{tabular}{lccc}
    \toprule
    \multirow{3}{*}{Correlation aggregation} & \multicolumn{3}{c}{SPair-71k (s)} \\
    & \multicolumn{3}{c}{@$\alpha_{\text{bbox}}$}  \\
    \cmidrule(r){2-4}
    & 0.05 & 0.10 & 0.15\\
    \midrule
    4D Conv$_{k=1}$ & 59.2 & 78.4 & 85.3 \\
    4D Conv$_{k=3}$  & 59.2 & 78.2 & 85.2 \\
    4D Conv$_{k=5}$  & 39.2 & 67.9 & 79.0 \\
    \midrule
    FastFormer~\citeyearpar{kim2022transformatcher, wu2021fastformer}  & 59.5 & 78.9 & 85.7 \\
    \midrule
    PlainMamba~\citeyearpar{yang2024plainmamba} & 56.7 & 78.5 & 85.7 \\
    Mamba$_\textrm{4D}$ & 59.3 & 78.8 & 85.6 \\
    + bidirectional & 59.0 & 78.6 & 85.5 \\
    + Z-order curve~\citeyearpar{morton1966computer} & 58.4 & 79.0 & 85.6 \\
    + ascending order & 58.4 & 78.2 & 85.2 \\
    + descending order & \textbf{59.9} & \textbf{79.3} & \textbf{86.2} \\
    \bottomrule
  \end{tabular}%
  \label{tbl:correlation_aggregation_ablation}

\end{minipage}
\vspace{-2.5mm}
\end{table*}

\noindent
\textbf{Evaluation metric.}
We use the Percentage of Correct Keypoints (PCK), which is the standard evaluation metric for semantic correspondence. Given $M$ predicted and ground-truth target keypoint pairs $\mathcal{K}=\{(\hat{\mathbf{k}}^{(m)}_t, \mathbf{k}^{(m)}_t)\}_{m=1}^M$, and a tolerance factor $\alpha_\tau$, PCK is measured by:
\begin{equation}
\textrm{PCK}(\mathcal{K}) = \frac{1}{M} \sum_{m=1}^M \mathbbm{1} \left[ \| \hat{\mathbf{k}}^{(m)}_t - \mathbf{k}^{(m)}_t \| \leq \alpha_\tau \cdot \max(w_\tau, h_\tau) \right],
\end{equation}
where $w_\tau$ and $h_\tau$ are the width and height of either the image or the object bounding box.

\subsection{Performance on Semantic Matching}
We evaluate \methodName{} on the PF-PASCAL~\citep{ham2017proposal} and SPair-71k~\citep{min2019spair} datasets. 
The results are shown in Table~\ref{tbl:pfpascal_spair}, where \methodName{} outperforms existing methods on both datasets, without relying on particularly large image sizes or computationally expensive backbones like Stable Diffusion. Moreover, our method incurs a reasonable computational overhead in terms of latency and memory usage.

Fig.~\ref{fig:mamba_viz} visualizes the effect of our proposed similarity-aware selective scan. We observe that our final refined correlation map $\hat{\mathbf{C}}$ better localizes keypoints, as indicated by well-localized regions with high similarity scores.
The initial correlation map prior to the similarity-aware selective scan shows high similarities at ground-truth locations, validating our choice of using the scores from the final correlation map to sort the multi-level correlation map.
\subsection{Analysis on the facet used}
\label{subsec:facet_analysis}

When using DINOv2, we can utilize different facets: key, query, value, or token. In Table~\ref{tbl:feature_facet_ablation}, we evaluate the PCK on the 'small' subset of SPair-71k when using each facet from the final layer (layer 11) of the DINOv2 backbone to establish a single-layer correlation map. We observe that the performance increases in the order of key, query, value, and token, demonstrating that the output token features are most effective for establishing semantic correspondences.

When using multi-level features (layers 4-11), we further experiment with incorporating additional features from other facets to improve performance. We default the multi-level correlation aggregation to a linear projection to a single-layer correlation map. The results in Table~\ref{tbl:feature_facet_multi_ablation} show that using all feature sources results in the best performance. However, the most significant performance increase occurs when additionally using the value features; we observe a 1.9\% increase at the 0.10 threshold compared to only a 0.1\% increase when adding other facets. To balance performance and computational overhead, we choose to use token and value features, resulting in $2L$ layers of features when extracting features across $L$ layers.

\subsection{Analysis on feature and correlation aggregation}
\label{subsec:aggregation_visualization}

\noindent
\textbf{Feature aggregation analysis.} Table~\ref{tbl:feature_aggregation_ablation} presents the comparative performance when using different feature aggregation schemes, evaluated on the \textit{small} subset of SPair-71k. As we extract multi-level features and compute multi-level correlation maps, we obtain a single-level refined correlation map after feature aggregation using a single $1 \times 1$ convolution layer in these experiments. Using no feature aggregation ('None') defaults to using the extracted features directly.
When applying Mamba$_\textrm{2D}$, we flatten the multi-level feature map $\mathbf{F} \in \mathbb{R}^{2L \times H \times W}$ to a sequence in $\mathbb{R}^{2L \times (H \times W)}$, which is then input to a Mamba layer. We also experiment with bi-directional selective scans, considering that an image does not have a fixed beginning or end, unlike a temporal sequence.
The Z-order curve~\citep{morton1966computer} is a space-filling curve that passes through every point in a high-dimensional discrete space while preserving spatial proximity, which has been effective in prior work~\citep{wu2023point, liang2024pointmamba} and is applicable to the scan order of Mamba. We find that using a series of 2D convolutional layers with a kernel size of $5$ performs the best - but considering that the performance improvement is marginal despite the  high increase in FLOPs and parameter overhead, we use a kernel size of $3$ in our experiments.

\noindent
\textbf{Correlation aggregation analysis.} Table~\ref{tbl:correlation_aggregation_ablation} presents the comparative performance when using different correlation aggregation schemes, evaluated on the \textit{small} subset of SPair-71k. Based on the results of the feature aggregation comparison (Table~\ref{tbl:feature_aggregation_ablation}), we default the feature aggregation scheme to 2D Conv$_{k=3}$ for these experiments.
Applying a vanilla transformer~\citep{dosovitskiy2020image} to the correlation tensor results in out-of-memory errors even on a single batch on an RTX 3090 GPU; therefore, we opt for FastFormer~\citep{wu2021fastformer} as the linear-complexity attention-based correlation aggregation scheme~\citep{kim2022transformatcher}. 
When applying Mamba$_\textrm{4D}$, we flatten the multi-level correlation map $\mathbf{C} \in \mathbb{R}^{L \times H \times W \times H \times W}$ to a sequence in $\mathbb{R}^{L \times (H \times W \times H \times W)}$, which is then input to a Mamba layer.
'Ascending order' and 'Descending order' indicate that the flattened correlation tensor is sorted in either ascending or descending order based on the similarity scores from the final correlation map of the multi-level correlation map. Among different schemes, we find that processing a descending-order sorted multi-level correlation map shows to be the best alternative to induce similarity-awareness in the selective scan algorithm, verifying the design choices of \methodName{}.
Fig.~\ref{fig:main_qual} compares different correlation aggregation schemes, showing that \methodName{} establishes more robust and accurate semantic correspondences under large viewpoint or scale variations. We provide qualitative comparisons between different selective scanning schemes in Fig.~\ref{fig:scan_comparison}.
We include further analyses and comparisons in Appendix~\ref{sec:feature_backbone_comparison},~\ref{sec:statistical_significance},~\ref{sec:appendix_flops},~\ref{sec:appendix_larger_resolutions}.

\section{Failure case analysis}
\label{sec:appendix_failure_case}

We include qualitative examples of failure cases of our method in Figure~\ref{fig:failure_case}. Firstly, \methodName{} tends to fail in scenarios involving large viewpoint variations combined with symmetry ambiguity, where our model struggles to accurately distinguish positions in symmetric instances. Secondly, \methodName{} may not follow the ground truth in the presence of multiple instances and semantic ambiguity.
We suggest that incorrect correspondence predictions due to semantic ambiguity could still be considered as semantic correspondences in a broader sense. This opens up interesting future directions, such as exploring many-to-many semantic correspondences instead of just one-to-one.

\section{Conclusion}
\label{sec:conclusion}

We introduced \methodName, a novel approach for semantic correspondence that models the high-dimensional correlation space using selective state-space models, treating multi-level correlation scores as states. By leveraging the efficiency of Mamba's linear-complexity algorithm and implementing a similarity-aware selective scan mechanism, \methodName effectively refines 4D correlation tensors without compromising feature map resolution or receptive field. Our evaluations on standard benchmarks demonstrate that \methodName outperforms existing methods while maintaining computational efficiency. This work not only advances the state of the art in semantic correspondence but also highlights the potential of applying SSMs to high-dimensional data, encouraging further exploration into integrating feature-metric and correlation-metric approaches.

\clearpage
\noindent
\small
\textbf{Acknowledgement.} 
Seungwook Kim was supported by the Hyundai-Motor Chung Mong-koo Foundation.
This work was also supported by the NRF grant (RS-2021-NR059830 (50\%)) and the IITP grants (RS-2022-II220113: Developing a Sustainable Collaborative Multi-modal Lifelong Learning Framework (45\%), RS-2019-II191906: AI Graduate School Program at POSTECH (5\%)) funded by the Korea government (MSIT).

{
    \small
    \bibliographystyle{ieeenat_fullname}
    \bibliography{01_main}
}

\clearpage
\newpage
\appendix
\onecolumn

\section{Additional implementation details}
\label{sec:appendix_implementation_details}

During training of \methodName in Tab.~\ref{tbl:pfpascal_spair}, we use an effective batch size of 80 by distributing 10 batches to 8 RTX 3090 GPUs.
For other comparison and ablative experiments, we run the experiments on a 'small' subset of SPair-71k, which is around 20\% the size of the original SPair-71k dataset~\citep{min2019spair}, with varying effective sizes across 2 GPUs.
The batch sizes vary because different feature and correlation aggregation schemes required different amount of vRAM.
For example, when using FastFormer~\citep{wu2021fastformer}, only 3 batches could fit into a single GPU when training.

\smallbreak
\noindent
\textbf{Details of soft sampler (Sec.~\ref{subsec:keypoint_transfer}.}
Given a source keypoint $\mathbf{k}_s = (x_{k_s}, y_{k_s})$, we define a soft sampler $\mathbf{W}^{\mathbf{k_s}} \in \mathbbm{R}^{H\times W}$:
\begin{equation}
\mathbf{W}^{\mathbf{k_s}}(i,j) = \frac{\textrm{max}(0,\tau - \sqrt{(x_{k_s} - j)^2 + (y_{k_s} - i)^2})}{\sum_{i'j'}\textrm{max}(0,\tau - \sqrt{(x_{k_s} - j')^2 + (y_{k_s} - i')^2})},
\end{equation}
where $\tau$ is a distance threshold from the keypoint, and $\sum_{ij}\mathbf{W}^{k_s}(i,j)=1$.
The role of the soft sampler is to sample each transferred keypoint $\mathbf{\hat{P}}(i,j)$ by assigning weights which are inversely proportional to the distance to the keypoint $\mathbf{k}_s.$
We can obtain sub-pixel accurate keypoint matches as follows:
\begin{equation}
\mathbf{\hat{k}}_t = \sum_{(i,j)\in H\times W} \mathbf{\hat{P}}(i,j)\mathbf{W}^{k_s}(i,j).
\end{equation}
We use $\tau=0.1$ for training, and $\tau=0.05$ for inference.

\smallbreak
\noindent
\textbf{Experimental environment.}
All experiments are run on a machine with an Intel(R) Xeon(R) Gold 6242 CPU, with up to 8 GeForce RTX 3090 GPUs.

\section{Additional details of baseline methods}
\label{sec:appendix_baseline_details}

We provide the details of each baseline approach (shown in Table~\ref{tbl:pfpascal_spair} of the main manuscript) in Table~\ref{tbl:baseline_details}, which was omitted due to spatial constraints.

\begin{table*}[h]
  \caption{\textbf{Additional details of baseline methods.}}
  \label{tbl:baseline_details}
  \centering
  \resizebox{1.0\textwidth}{!}{%
  \begin{tabular}{lccc}
    \toprule
    
    Method & Feature backbone & Supervision & Data augmentation \\ 
    \midrule
    DHPF, CHM, MMNet, PWarpC-NCNet, NeMF, SCorrSAN  & ResNet101 & kp-pair & x \\
    TransforMatcher, CATs++, HCCNet, UFC & ResNet101 & kp-pair & o \\
    DIFT & SD2.1 & None & x \\ 
    DINO + SD$_\textrm{zero-shot}$ & DINOv2, SD1.5 & None & x \\
    DINO + SD$_\textrm{supervised}$ & DINOv2, SD1.5 & kp-pair & x \\    
    Diffusion Hyperfeatures & SD1.5 & None & x \\ 
    \citet{hedlin2024unsupervised} & SD1.4 & None & x \\ 
    SD4Match & SD2.1 & kp-pair & x \\
    \methodName & DINOv2 & kp-pair & o \\
    \bottomrule

  \end{tabular}%
  }
\end{table*}

\section{Effect of positional encoding}
\label{sec:appendix_positional_encoding}

Currently, we do not explicitly encode the potential spatial relationships between correlation elements during the sorting and processing steps.
While spatial relationships are important, our primary goal is to resolve ambiguities by focusing on the most significant correspondences first. 
Prioritizing high-similarity scores enables the model to establish a strong contextual foundation.
Nonetheless, we experiment the effect of fixed 4D sinusoidal encoding and learnable positional embedding to further analyze the effect of \textit{explicitly} encoding spatial relationships in Table~\ref{tbl:appendix_positional_encoding}, where it can be seen that there is no clear visible improvement in performance with the integration of sinusoidal or learned positional embeddings.
Investigating more sophisticated methods to integrate spatial context could potentially enhance the model's ability to capture inter-image relationships.

\begin{table}[h]
  \caption{\textbf{Effect of using 4D sinusoidal positional encoding.}}
  \label{tbl:appendix_positional_encoding}
  \centering
  \resizebox{0.5\textwidth}{!}{%
  \begin{tabular}{lccc}
    \toprule
    
    Method & PCK @ 0.05 & PCK @ 0.10 & PCK @ 0.15 \\ 
    \midrule
    Ours & 61.6 & 77.8 & 84.3 \\ 
    Ours with Sinusoidal P.E. & 61.2 & 77.8 & 84.2 \\
    Ours with Learnable P.E. & 61.5 & 77.6 & 84.5 \\
    \bottomrule

  \end{tabular}%
  }
\end{table}

\section{Feature backbone / Data augmentation comparison}
\label{sec:feature_backbone_comparison}

\begin{table}[h]
  \caption{\textbf{PCK of \methodName on SPair-71k when using varying feature backbones and data augmentation.} We follow the data augmentation scheme used in CATs~\citep{cho2021cats} and TransforMatcher~\citep{kim2022transformatcher}}
  \label{tbl:varying_backbone_aug}
  \centering
  \resizebox{0.5\textwidth}{!}{%
  \begin{tabular}{lcccc}
    \toprule
    
    Backbone & Data aug. & PCK@0.05 & PCK@0.10 & PCK@0.15 \\ 
    \midrule
    ResNet101 & x & 38.2 & 53.3 & 61.3 \\
    ResNet101 & o & 41.0 & 58.5 & 67.4 \\
    DINOv2 & x & 57.9 & 74.6 & 81.8 \\
    DINOv2 & o & 61.6 & 77.8 & 84.3 \\ 
    \bottomrule
  \end{tabular}%
  }
\end{table}

We provide the results of \methodName when using varying backbones, with or without data augmentation, on SPair-71k for a fairer comparison in Table~\ref{tbl:varying_backbone_aug}.
Noting that PCK@0.05/0.10 for TransforMatcher~\citep{kim2022transformatcher} are 32.4/53.7 with data augmentation, these results show that the similarity-aware selective scan shows enhanced efficacy over multiple layers of additive attention (FastFormers~\citep{wu2021fastformer}).

\section{Statistical significance of performance gap in comparison to FastFormers}
\label{sec:statistical_significance}

We conduct 3 repeated experiments with varying seeds to report the mean and variance of PCK results on the 'small' subset of SPair-71k in Table~\ref{tbl:statistical_significance}.
While the performance gain is not dramatic, \methodName offers advantages in terms of computational overhead (memory, latency) as previously shown in Table~\ref{tbl:memory_usage_and_latency}.

\begin{table*}[h]
  \caption{\textbf{PCK results on SPair-71K over multiple runs} We report the results when using FastFormers in comparison to our similarity-aware selective scan as the correlation aggregation. The experiments were conducted 3 times - the mean and standard variation across the runs are reported. It can be seen that our scheme consistently yields better performances across PCK thresholds.}
  \label{tbl:statistical_significance}
  \centering
  \resizebox{0.75\textwidth}{!}{%
  \begin{tabular}{lccc}
    \toprule
    
    Method & PCK@0.05 & PCK@0.10 & PCK@0.15 \\ 
    \midrule
    FastFormers~\citep{kim2022transformatcher} (6 layers) & $59.9\pm0.74$ & $76.9\pm1.40$ & $83.9\pm1.27$\\
    Mamba + Similarity-aware Selective Scan (Ours) & $60.6\pm0.54$ & $78.2\pm0.76$ & $85.0\pm0.86$\\
    \bottomrule

  \end{tabular}%
  }
\end{table*}

\section{Analysis on efficiency of \methodName}
\label{sec:efficiency_analysis}

For an intuitive overview, we measure module-wise maximum GPU memory usage and latency in Table~\ref{tbl:memory_usage_and_latency}. The values are cumulative in the order of DINOv2 (feature extraction), feature aggregation, and correlation aggregation. This shows that our design incurs the lowest latency while using less memory and fewer parameters than FastFormer, demonstrating a favorable balance between computational overhead and performance\footnote{We report the FLOPs of \methodName{} in Appendix~\ref{sec:appendix_flops}.}. 

\begin{table*}[h]
\vspace{-3.0mm}
  \caption{\textbf{Memory, Latency and \# Params comparison across correlation schemes.} Our scheme strikes the most favorable balance between performance and efficiency.}
  \label{tbl:memory_usage_and_latency}
  \centering
  \resizebox{0.9\textwidth}{!}{%
  \begin{tabular}{lccc}
    \toprule
    
    Module & GPU Memory (GB) & Latency (ms) & \# Params \\ 
    \midrule
    DINOv2~\citep{oquab2023dinov2} & 0.97 & 10.3 & 86.6M \\
    Feature aggregation & 1.17 & 12.3 & 42.5M \\
    \midrule
    Correlation aggregation &&& \\
    - Conv4D$_\textrm{k=3}$~\citep{min2021convolutional} & 1.17 & 41.0 & 1.3K \\
    - FastFormers~\citep{kim2022transformatcher} (6 layers) & 1.67 & 28.8 & 26.0K \\
    - Mamba$_{4D}$ + Similarity-aware Selective Scan (Ours) & 1.64 & 16.4 & 5.1K \\
    \bottomrule

  \end{tabular}%
  }
\end{table*}

\section{FLOPs analysis of \methodName}
\label{sec:appendix_flops}

In the Table~\ref{tbl:ptflops_calflops}, we report the FLOPs of \methodName using open-source libraries \texttt{ptflops} and \texttt{calflops}.

\begin{table*}[h]
  \caption{\textbf{FLOPs of \methodName measured using open-source libraries.}}
  \label{tbl:ptflops_calflops}
  \centering
  \resizebox{0.6\textwidth}{!}{%
  \begin{tabular}{lcc}
    \toprule
    
    Module & \texttt{ptflops} & \texttt{calflops} \\ 
    \midrule
    DINOv2~\citep{oquab2023dinov2} & 359.32G & 358.99G \\
    Feat. agg & 2.45T & 2.45 T \\
    \midrule
    Conv4D$_{k=3}$~\citep{min2021convolutional} & 2.06G & 2.06G \\
    FastFormers~\citep{kim2022transformatcher} (6 layers) & 43.54G & 43.05G \\
    Mamba + Similarity-aware Selective Scan (Ours) & 27.54M & 3.84G \\
    \bottomrule

  \end{tabular}%
  }
\end{table*}

While FLOPs serve as a standardized measure of computational complexity, we noticed that existing libraries fail to accurately capture the FLOPs of various modules due to technical complexities, \textit{e.g.,} reliance on operations registered as \texttt{nn.Modules}. Additionally, certain libraries for measuring FLOPs crash when encountered with hardware-optimized algorithms from \texttt{xFormers}~\citep{xFormers2022}, which are used in the DINOv2 backbone of our method.
Consequently, we believe that this measurement may not be entirely fair or representative of the actual computational overhead and efficiency.

To address this gap, we conduct a theoretical calculation of FLOPs for varying correlation aggregation schemes. We consider an input with dimensions $N \times C = 30^4 \times 16$, consistent with \methodName.
We assume the same dimensions for the input and output \textit{i.e.,} $C = C_\textrm{in} = C_\textrm{out}$.

\smallbreak
\noindent
\textbf{4D convolution, kernel size 3}. \\
$2 \times N \times C_\textrm{in} \times C_\textrm{out} \times k^4 = 33.6$ GFLOPs

\smallbreak
\noindent
\textbf{Vanilla dot-product attention}.
Assuming single head, QKV dim = 16. \\
QKV projection: $3 \times (2 \times N \times C_\textrm{in} \times C_\textrm{out})$ \\
Dot-product: $2 \times (N^2 \times C)$ \\
Softmax: $3 \times (N^2)$ \\
Weighted sum of V: $2 \times (N^2) \times C$ \\
Total = 44.0 TFLOPs

\smallbreak
\noindent
\textbf{FastFormers (Additive attention)}. Assuming single head, QKV dim = 16.\\
QKV projection: $3 \times (2 \times N \times C_\textrm{in} \times C_\textrm{out})$ \\
Softmax and weighted sum: $2 \times (3 \times N + 2 \times N \times C)$ \\ 
Global vector addition: $2 \times (N \times C)$ \\ 
Projection: $2 \times N \times C^2$
Total = 1.74 GFLOPs

\smallbreak
\noindent
\textbf{Mamba: selective state-space machines}. Hyperparameters following \methodName. \\
Input projection: $2 \times 2 \times N \times C_\textrm{in} \times C_\textrm{inner}$ \\ 
1D convolution: $2 \times C_\textrm{inner} \times k \times C_\textrm{inner}$ \\
Projection to A, B, dt: $2 \times N \times C_\textrm{inner} times (2 \times d_\textrm{model} + 1)$ \\ 
Selective scan: $ 9 \times N \times d_\textrm{model} \times d_\textrm{state}$ \\
Element-wise multiplication: $N \times C_\textrm{inner}$ \\
Output projection: $2 \times N \times C_\textrm{inner} \ times C_\textrm{in}$ \\ 
Total FLOPs = 23.1 GFLOPs

\smallbreak
\noindent
\textbf{Ours: Selective state-space machines with Similarity-aware Selective Scan}. Same as above, but additional sorting overhead. Assuming each comparison and swap operation involves approximately 4 FLOPs: \\
Sorting: $4 \times (N log N) = 0.064$GFLOPs \\ 
Total FLOPs = 23.2 GFLOPs

Note that the above values ignore many details, including activation, normalization, residual connections, or actual number of aggregation layers used.
The above theoretical calculation serve to provide a vague estimate of FLOPs for each scheme.
However, we suggest that the number of FLOPs does not directly translate to computational overhead in learning-based methods, as many variables such as parallelism, hardware optimization, and intermediate representations directly impact GPU memory usage and latency.

\section{Generalizability of \methodName}
\label{sec:appendix_pfwillow}

\smallbreak
\noindent
\textbf{Trained on PF-PASCAL, evaluated on PF-WILLOW.}
We present the results of \methodName{} on the PF-WILLOW~\citep{ham2017proposal} dataset.
The PF-WILLOW dataset contains 900 image pairs for testing only and is evaluated using the model trained on the PF-PASCAL dataset.
The results are illustrated in Table~\ref{tbl:pfwillow}, where it can be seen that while \methodName{} performs competitively, it does not outperform existing methods.
This is unlike our results on the PF-PASCAL and SPair-71k datasets (Table~\ref{tbl:pfpascal_spair}), where \methodName{} outperforms all existing benchmarks.
This may be attributed to supervised training, which causes the feature and correlation aggregation layers to be trained specifically for the training domain.
Another possibility is that the Mamba layer lacks generalizability to unseen domains compared to other methods built on convolutional or attention-based layers.

\begin{table}[h]
  \caption{\textbf{Results of \methodName on the PF-WILLOW dataset.} We perform competitively with existing methods, but do not outperform all existing methods unlike on PF-PASCAL or SPair-71k.}
  \label{tbl:pfwillow}
  \centering
  \resizebox{0.5\textwidth}{!}{%
  \begin{tabular}{lcccc}
    \toprule
    \multirow{3}{*}{Method} & \multicolumn{4}{c}{PF-WILLOW} \\
    & \multicolumn{2}{c}{@$\alpha_{\text{bbox}}$} & \multicolumn{2}{c}{@$\alpha_{\text{bbox-kp}}$} \\
    \cmidrule(r){2-3}
    \cmidrule(r){4-5}
    & 0.05 & 0.10 & 0.05 & 0.10 \\
    \midrule
    DHPF~\citeyearpar{min2020learning} & 49.5 & 77.6 & - & 71.0 \\
    CHM~\citeyearpar{min2021convolutional} & 52.7 & 79.4 & - & 69.6 \\
    CATs++~\citeyearpar{cho2022catspp} & 56.7 & 81.2 & 47.0 & 72.6 \\
    PWarpC-NCNet~\citeyearpar{truong2022probabilistic} & - & - & 48.0 & 76.2 \\
    TransforMatcher~\citeyearpar{kim2022transformatcher} & - & 76.0 & - & 65.3 \\
    NeMF~\citeyearpar{hong2022neural} & - & - & 60.8 & 75.0\\
    SCorrSAN~\citeyearpar{huang2022learning} & 54.1 & 80.0 & - & - \\
    HCCNet~\citeyearpar{kim2024hccnet} & - & 74.5 & - & 65.5 \\
    UFC~\citeyearpar{hong2023unifying} & 58.6 & 81.2 & 50.4 & 74.2 \\
    DIFT~\citeyearpar{tang2023emergent} & 58.1 & 81.2 & 44.8 & 68.0 \\
    DINO+SD$_\textrm{zero-shot}$~\citeyearpar{zhang2024tale} & - & - & - & - \\
    DINO+SD$_\textrm{sup}$~\citeyearpar{zhang2024tale} & - & - & - & - \\
    Diffusion Hyperfeatures~\citeyearpar{luo2024diffusion} & - & 78.0 & - & - \\
    ~\citet{hedlin2024unsupervised} & 53.0 & 84.3 & - & - \\
    SD4Match~\citeyearpar{li2023sd4match} & - & - & 52.1 & 80.4 \\
    \midrule
    Ours & 56.2 & 81.1 & 47.4 & 72.1  \\
    \bottomrule
  \end{tabular}%
  }
\end{table}

\smallbreak
\noindent
\textbf{Trained on SPair-71k, evaluated on PF-PASCAL.}
While we provide the generalization performance of \methodName{} on the PF-WILLOW dataset in Table~\ref{tbl:pfwillow}, we report additional generalization results in Table~\ref{tbl:pfpascal_to_spair}.
Results on PF-PASCAL were trained on SPair-71k, and vice versa.
The results indicate that while the generalizability of \methodName{} is not state-of-the-art, it generalizes competitively with other state-of-the-art methods in certain cases, such as being trained on PF-PASCAL and tested on SPair-71k.
While domain generalization is advantageous, we suggest that a lack of cross-dataset generalization does not diminish the overall significance of our method.
If large-scale datasets for semantic correspondence become available, this problem is likely to be alleviated significantly for all semantic matching methods.

\begin{table*}[h]
  \caption{\textbf{PCK on SPair-71k after being trained on PF-PASCAL.}}
  \label{tbl:pfpascal_to_spair}
  \centering
  \resizebox{0.7\textwidth}{!}{%
  \begin{tabular}{lccc}
    \toprule
    
    Model & PCK@0.05 & PCK@0.10 & PCK@0.15 \\ 
    \midrule
    CATs~\citep{cho2021cats} & 13.6 & 27.0 & - \\
    TransforMatcher~\citep{kim2022transformatcher} & - & 30.1 & - \\
    SD4Match~\citep{li2023sd4match} & 27.2 & 40.9 & - \\
    \methodName (Ours) & 26.5 & 40.9 & 49.1 \\
    \bottomrule

  \end{tabular}%
  }
\end{table*}

\section{Comparison on the DINOv2 layers used}
\label{sec:appendix_layer}

We show the comparative experiments on the layers if DINOv2 used in this work to validate our use of layers 4-11. 
The experiments were carried out on the 'small' set of SPair-71k.
The results in Tab.~\ref{tbl:feature_layer_ablation} shows that better features can be obtained across the depths of the DINOv2 backbone, with the 11th layer token features exhibiting the best performance.
Tab.~\ref{tbl:feature_layer_multi_ablation} aims to choose the best combination of layers to extract the feature maps from.
While the PCK performance improves gracefully as more layers are used, we choose to use layers 4-11 as the performance improvement beyond that becomes diminishing, and using layers 4-11 provides us with a favorable compromise between memory usage (around 70\% memory usage compared to using all 0-11 layers) and PCK performance.

\begin{table}[h]
\centering
\begin{minipage}{0.48\textwidth}
\centering
      \caption{\textbf{Comparison between different layers of the DINOv2 backbone.}}
  \centering
\begin{tabular}{cccc}
    \toprule
    \multirow{3}{*}{Layers used} & \multicolumn{3}{c}{SPair-71k (s)} \\
    & \multicolumn{3}{c}{@$\alpha_{\text{img}}$}  \\
    \cmidrule(r){2-4}
    &  0.05 & 0.10 & 0.15\\

    \midrule
     0 & 0.9 & 3.8 & 8.2 \\
     1 & 1.5 & 5.3 & 11.2 \\
     2 & 1.7 & 6.1 & 12.2 \\
     3 & 4.2 & 11.1 & 18.8 \\
     4 & 7.3 & 16.1 & 24.6 \\
     5 & 10.2 & 20.6 & 29.8 \\
     6 & 13.1 & 23.6 & 31.8 \\
     7 & 17.5 & 29.8 & 39.0 \\
     8 & 20.7 & 35.2 & 45.7 \\
     9 & 23.9 & 40.3 & 51.5 \\
     10 & 25.2 & 42.5 & 54.1 \\
     11 & 25.2 & 43.1 & 55.7 \\
    \bottomrule
  \end{tabular}%
\label{tbl:feature_layer_ablation}

\end{minipage}%
\hfill%
\begin{minipage}{0.48\textwidth}
\centering
     \caption{\textbf{Comparison between different layers combinations of the DINOv2 backbone}.}
  \centering
  \begin{tabular}{cccc}
    \toprule
    \multirow{3}{*}{Layers used} & \multicolumn{3}{c}{SPair-71k (s)} \\
    & \multicolumn{3}{c}{@$\alpha_{\text{img}}$}  \\
    \cmidrule(r){2-4}
    &  0.05 & 0.10 & 0.15\\

    \midrule
     11 & 25.2 & 43.1 & 55.7 \\
     10-11 & 29.2 & 46.4 & 56.8 \\
     9-11 & 28.9 & 46.7 & 58.0 \\
     8-11 & 29.6 & 47.4 & 58.3 \\
     7-11 & 30.4 & 48.5 & 58.8 \\
     6-11 & 30.8 & 48.4 & 58.7 \\
     5-11 & 30.9 & 48.4 & 58.6 \\
     4-11 & 30.8 & 48.6 & 59.0 \\
     3-11 & 31.0 & 48.7 & 59.0 \\
     2-11 & 31.2 & 48.9 & 58.7 \\
     1-11 & 31.4 & 48.9 & 58.8 \\
     0-11 & 31.4 & 48.9 & 58.7 \\
    \bottomrule
  \end{tabular}%
  \label{tbl:feature_layer_multi_ablation}

\end{minipage}
\end{table}

\section{Comparison on the layers used for sorting}
\label{sec:sorting_layer}

Currently, when performing multi-level correlation aggregation via the Similarity-Aware Selective Scan, we sort the correlation sequence based on the similarity scores from the final correlation map (the 2$L$-th level,~\cref{subsec:correlation_computation_aggregation}).
We compare the results when using a different standard for sorting the correlation sequence in~\cref{tbl:sorting_comparison}, where we show that our current configuration of using the 2$L$-th level demonstrates the best results.

\begin{table}[h]
  \caption{\textbf{Effect of using different configurations for sorting the correlation sequence.}}
  \label{tbl:sorting_comparison}
  \centering
  \resizebox{0.5\textwidth}{!}{%
  \begin{tabular}{lccc}
    \toprule
    
    Method & PCK @ 0.05 & PCK @ 0.10 & PCK @ 0.15 \\ 
    \midrule
    Last layer (2$L$-th, Ours) & 61.6 & 77.8 & 84.3 \\ 
    Penultimate-layer (2$L$-1th) & 61.0 & 77.7 & 84.2 \\
    Mean across layers & 60.9 & 76.6 & 83.8 \\
    \bottomrule

  \end{tabular}%
  }
\end{table}

\section{PCK per image v.s. PCK per point}
\label{sec:appendix_pck_per_point_image}

While it is conventional to calculate the mean PCK per image (sum of image-wise PCK averaged over the number of images) when reporting the PCK results, some methods confuse this concept with PCK per point (sum of pair-wise PCK averaged over the number of point pairs).
Tab.~\ref{tbl:pfpascal_spair_appendix} shows the results, where it can be seen that \methodName evaluated using PCK-per-point (denoted as MambaMatcher*) yields higher values in comparison.

\begin{table*}
  \caption{\textbf{Results of \methodName on PF-PASCAL and SPair-71k datasets.} \methodName outperforms existing baselines on both datasets. 
   \methodName* denotes PCK-per-point metrics, which outperforms \methodName.
  This shows that PCK-per-point yields higher results in comparison to PCK-per-image.}
  \label{tbl:pfpascal_spair_appendix}
  \centering
  \resizebox{\textwidth}{!}{%
  \begin{tabular}{llcccccccc}
    \toprule
    \multirow{3}{*}{Method} & \multirow{3}{*}{Image res.} & \multicolumn{3}{c}{PF-PASCAL} & \multicolumn{3}{c}{SPair-71k} & \multirow{3}{*}{\shortstack{time\\(\emph{ms})}} & \multirow{3}{*}{\shortstack{memory\\(GB)}} \\

    & & \multicolumn{3}{c}{@$\alpha_{\text{img}}$} & \multicolumn{3}{c}{@$\alpha_{\text{bbox}}$} \\
    \cmidrule(r){3-5}
    \cmidrule(r){6-8}
    & & 0.05 & 0.10  & 0.15 & 0.05 & 0.10 & 0.15 \\
    \midrule
    DHPF~\citeyearpar{min2020learning} & 240$\times$240 & 75.7 & 90.7 & 95.0 & 20.9 & 37.3 & 47.5 & 58 & 1.6 \\
    CHM~\citeyearpar{min2021convolutional} & 240$\times$240 & 80.1 & 91.6 & 94.9 & 27.2 & 46.3 & 57.5 & 54 & 1.6  \\
    MMNet~\citeyearpar{zhao2021multi} & 224$\times$320 & 77.6 & 89.1 & 94.3 & - & 40.9 & - & 86 & - \\
    PWarpC-NCNet~\citeyearpar{truong2022probabilistic} & 400$\times$400 & 79.2 & 92.1 & 95.6 & 31.6 & 52.0 & 61.8 & - & - \\
    TransforMatcher~\citeyearpar{kim2022transformatcher} & 240$\times$240 & 80.8 & 91.8 & - & 32.4 & 53.7 & - & 54 & 1.6 \\
    NeMF~\citeyearpar{hong2022neural} & 512$\times$512 & 80.6 & 93.6 & - & 34.2 & 53.6 & - & 8500 & 6.3 \\
    SCorrSAN~\citeyearpar{huang2022learning} & 256$\times$256 & 81.5 & 93.3 & - & - & 55.3 & - & 28 & 1.5  \\
    HCCNet~\citeyearpar{kim2024hccnet} & 240$\times$240 & 80.2 & 92.4 & - & 35.8 & 54.8 & - & 30 & 2.0 \\
    CATs++~\citeyearpar{cho2022catspp} & 512$\times$512 & 84.9 & 93.8 & 96.8 & 40.7 & 59.8 & 68.5 & - & - \\
    UFC~\citeyearpar{hong2023unifying} & 512$\times$512 & \textbf{88.0} & 94.8 & 97.9 & 48.5 & 64.4 & {72.1} & - & - \\
    DIFT~\citeyearpar{tang2023emergent} & 768$\times$768 & 69.4 & 84.6 & 88.1 & 39.7 & 52.9 & - & - & - \\
    DINO+SD$_\textrm{zero-shot}$~\citeyearpar{zhang2024tale} & 840$^2$ / 512$^2$ & 73.0 & 86.1 & 91.1 & - & 64.0 & - & - & - \\
    DINO+SD$_\textrm{sup}$~\citeyearpar{zhang2024tale} & 840$^2$ / 512$^2$ & 80.9 & 93.6 & 96.9 & - & 74.6 & - & - & -  \\
    Diffusion Hyperfeatures~\citeyearpar{luo2024diffusion} & 224$\times$224 & - & 86.7 & - & - & 64.6 & - & 6620 & - \\
    ~\citet{hedlin2024unsupervised} & 0.93$\times$ori. & - & - & - & 28.9 & 45.4 & - & 90k$<$ & - \\
    SD4Match~\citeyearpar{li2023sd4match} & 768$\times$768 & 84.4 & \underline{95.2} & \underline{97.5} & {59.5} & {75.5} & - & - & -\\
    \midrule
    \methodName (Ours) & 420$\times$420 & 87.3 & \underline{95.9} & \textbf{98.2} & \underline{61.6} & \underline{77.8} & \underline{84.3} & 74 & 2.1 \\
    \methodName* (Ours) & 420$\times$420 & \underline{87.6} & \textbf{96.0} & \textbf{98.2} & \textbf{63.3} & \textbf{79.2} & \textbf{85.6} & 74 & 2.1 \\
    \bottomrule
  \end{tabular}%
  }
\end{table*}

\section{PCK per category}
\label{sec:appendix_pck_per_category}
\begin{table*}[htbp]
\centering
\caption{\textbf{Category-wise PCK on the SPair-71k dataset}.}
\resizebox{1.0\textwidth}{!}{%
\begin{tabular}{lcccccccccccccccccccccc}
\toprule
\textbf{Method} & \textbf{Aero} & \textbf{Bike} & \textbf{Bird} & \textbf{Boat} & \textbf{Bottle} & \textbf{Bus} & \textbf{Car} & \textbf{Cat} & \textbf{Chair} & \textbf{Cow} & \textbf{Dog} & \textbf{Horse} & \textbf{Motor} & \textbf{Person} & \textbf{Plant} & \textbf{Sheep} & \textbf{Train} & \textbf{TV} & \textbf{All} \\
\midrule
DINOv2~\citeyearpar{oquab2023dinov2} & 69.9 & 58.9 & 86.8 & 36.9 & 43.4 & 42.6 & 39.3 & 70.2 & 37.5 & 69.0 & 63.7 & 68.9 & 55.1 & 65.0 & 33.3 & 57.8 & 51.2 & 31.2 & 53.9 \\
DIFT~\citeyearpar{tang2023emergent}& 61.2 & 53.2 & 79.5 & 31.2 & 45.3 & 39.8 & 33.3 & 77.8 & 34.7 & 70.1 & 51.5 & 57.2 & 50.6 & 41.4 & 51.9 & 46.0 & 67.6 & 59.5 & 52.9 \\
SD+DINO~\citeyearpar{zhang2024tale} & 71.4 & 59.1 & 87.3 & 38.1 & 51.3 & 43.3 & 40.2 & 77.2 & 42.3 & 75.4 & 63.2 & 68.8 & 56.0 & 66.1 & 52.8 & 59.4 & 63.0 & 55.1 & 59.3 \\
NCNet~\citeyearpar{rocco2018neighbourhood} & 17.9 & 12.2 & 32.1 & 11.7 & 29.0 & 19.9 & 16.1 & 39.2 & 9.9 & 23.9 & 18.8 & 15.7 & 17.4 & 15.9 & 14.8 & 9.6 & 24.2 & 31.1 & 20.1 \\
PMNC~\citeyearpar{lee2021patchmatch} & 54.1 & 35.9 & 74.9 & 36.5 & 42.1 & 48.8 & 40.0 & 72.6 & 21.1 & 67.6 & 58.1 & 50.5 & 40.1 & 54.1 & 43.3 & 35.7 & 74.5 & 59.9 & 50.4 \\
TransforMatcher~\citeyearpar{kim2022transformatcher} & 59.2 & 39.3 & 73.0 & 41.2 & 52.5 & 66.3 & 55.4 & 67.1 & 26.1 & 67.1 & 56.6 & 53.2 & 45.0 & 39.9 & 42.1 & 35.3 & 75.2 & 68.6 & 53.7 \\
SCorrSAN~\citeyearpar{huang2022learning} & 57.1 & 40.3 & 78.3 & 38.1 & 51.8 & 57.8 & 47.1 & 67.9 & 25.2 & 71.3 & 63.9 & 49.3 & 45.3 & 49.8 & 48.8 & 40.3 & 77.7 & 69.7 & 55.3 \\
SD4Match~\citeyearpar{li2023sd4match} & 75.3 & 67.4 & 85.7 & 64.7 & 62.9 & 86.6 & 76.5 & 82.6 & 64.8 & 86.7 & 73.0 & 78.9 & 70.9 & 78.3 & 66.8 & 64.8 & 91.5 & 86.6 & \underline{75.5} \\
\midrule
\methodName (Ours) & 82.9 & 61.0 & 91.9 & 61.0 & 62.7 & 89.9 & 83.8 & 89.9 & 60.6 & 86.7 & 81.2 & 81.6 & 73.7 & 79.5 & 70.0 & 71.5 & 93.0 & 86.4 & \textbf{77.8} \\
\bottomrule
\end{tabular}
}
\label{tab:categorywise_pck}
\end{table*}

We present the category-wise PCK in Tab.~\ref{tab:categorywise_pck}, where it can be seen that \methodName yields the best results overall.

\section{Potential when using larger resolutions}
\label{sec:appendix_larger_resolutions}

In Table~\ref{tbl:larger_image_resolutions}, we report the GPU memory / latency usage when using different correlation aggregation module at varying image resolutions (thus, varying feature and correlation map resolutions). Note that the memory usage is cumulative i.e., maximum GPU memory usage during the forward run.
It can be seen that our similarity-aware selective scan incurs consistently lower GPU memory usage and latency compared to FastFormers. Most notably, the difference in latency is dramatic; the hardware optimizations of Mamba enables the similarity-aware selective scan to be performed with only a small increase in latency even when the image sizes become significantly larger. This further justifies our usage of Mamba, given larger image inputs i.e., consequently, longer correlation sequences.

\begin{table*}[h]
  \caption{\textbf{Efficiency comparison when using larger image resolutions.}}
  \label{tbl:larger_image_resolutions}
  \centering
  \resizebox{0.6\textwidth}{!}{%
  \begin{tabular}{lcccc}
    \toprule
    
    Image res. & Feature res. & Correlation agg. & GPU memory (GB) & latency (ms) \\ 
    \midrule
    420 & 30 & Ours & 1.64 & 16.4 \\
    420 & 30 & FastFormer & 1.67 & 28.8 \\
    560 & 40 & Ours & 3.25 & 16.6 \\
    560 & 40 & FastFormer & 3.16 & 28.9 \\
    700 & 50 & Ours & 6.47 & 17.7 \\
    700 & 50 & FastFormer & 6.27 & 55.6 \\
    \bottomrule

  \end{tabular}%
  }
\end{table*}

We report the performance of \methodName on a different set of image resolutions in Table~\ref{tbl:larger_image_resolutions_pck}.
It shows that while using larger image resolutions does result in improved PCK results, there are diminishing returns as the image resolutions becomes larger.

\begin{table*}[h]
  \caption{\textbf{PCK results on SPair-71k when using varying image resolutions.}}
  \label{tbl:larger_image_resolutions_pck}
  \centering
  \resizebox{0.8\textwidth}{!}{%
  \begin{tabular}{lccccc}
    \toprule
    
    Image res. & Feature res. & Correlation res. & PCK @ 0.05 & PCK @ 0.10 & PCK @ 0.15 \\ 
    \midrule
    238 & $17^2$ & $17^4$ & 26.4 & 39.7 & 46.5 \\
    420 & $30^2$ & $30^4$ & 61.6 & 77.8 & 84.3 \\
    840 & $60^2$ & $60^4$ & 64.2 & 78.4 & 85.2 \\
    \bottomrule

  \end{tabular}%
  }
\end{table*}

\section{Additional visualizations of refined correlation map}
\label{sec:appendix_correlation_visualization}

\begin{figure*}[t]
    \centering
    \includegraphics[width=0.65\textwidth]{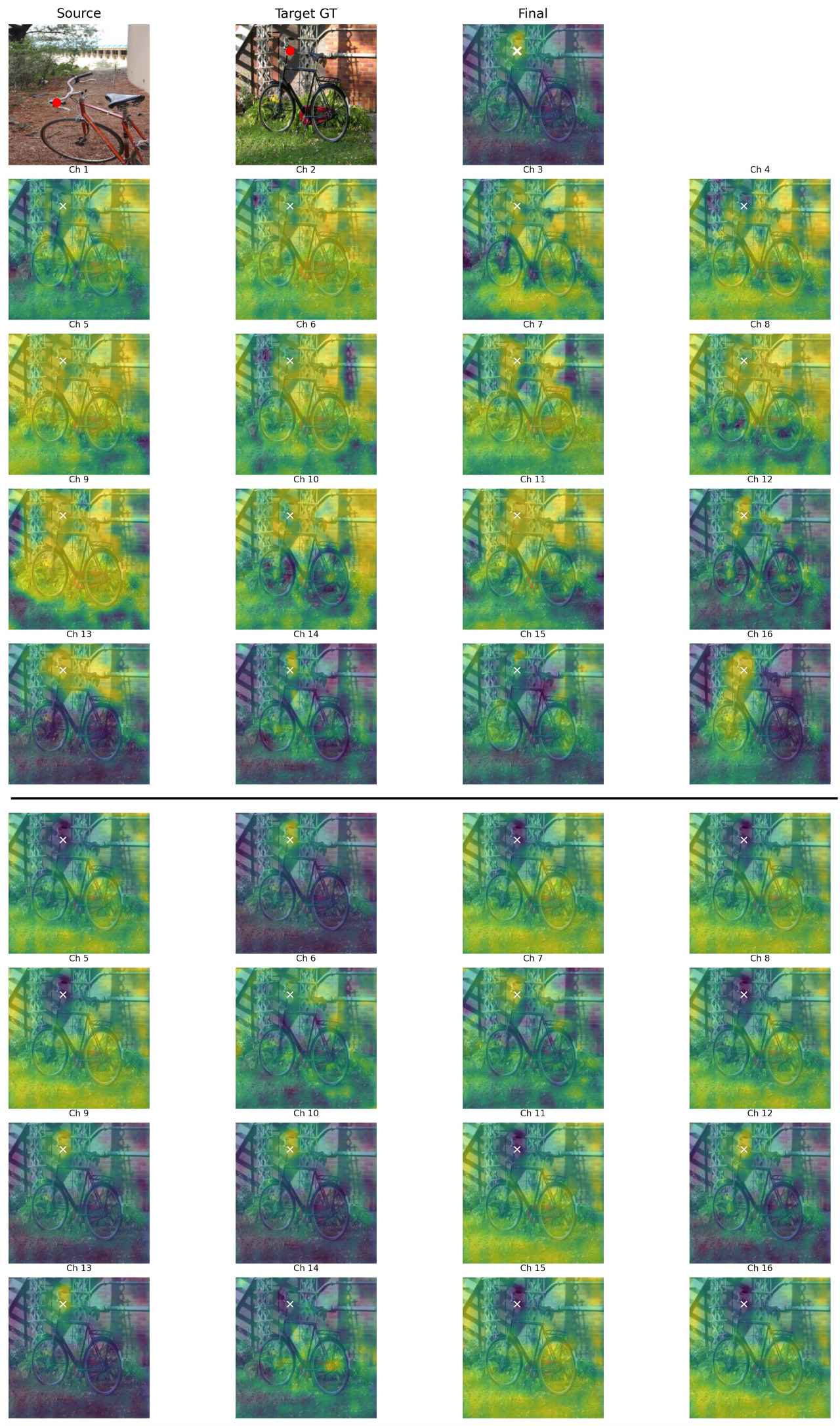}
    \vspace{-3mm}
    \caption{\textbf{Additional visualization of similarity-aware selective scan of \methodName}.
(First row) Source and target images with GT keypoints shown in red, and our final correlation tensor.
(First block of 4x4) The pre-refinement correlation maps, per channel. (Second block of 4x4) The post-refinement correlation maps, per channel.
As observed, after refinement, each correlation map is refined to be either \textit{perfectly} wrong, \ie, high attention everywhere other than the GT position, or accurately reflecting the keypoint position.
This visualization helps illustrate that during the final prediction of $\hat{\mathbf{C}}$ using linear projection, the wrong maps are effectively disregarded, and the accurate maps are primarily weighted for aggregation, resulting in our final accurate correlation map.}
    \label{fig:appendix_mamba_qual_1}
    \vspace{-3mm}
\end{figure*}

\begin{figure*}[t]
    \centering
    \includegraphics[width=0.65\textwidth]{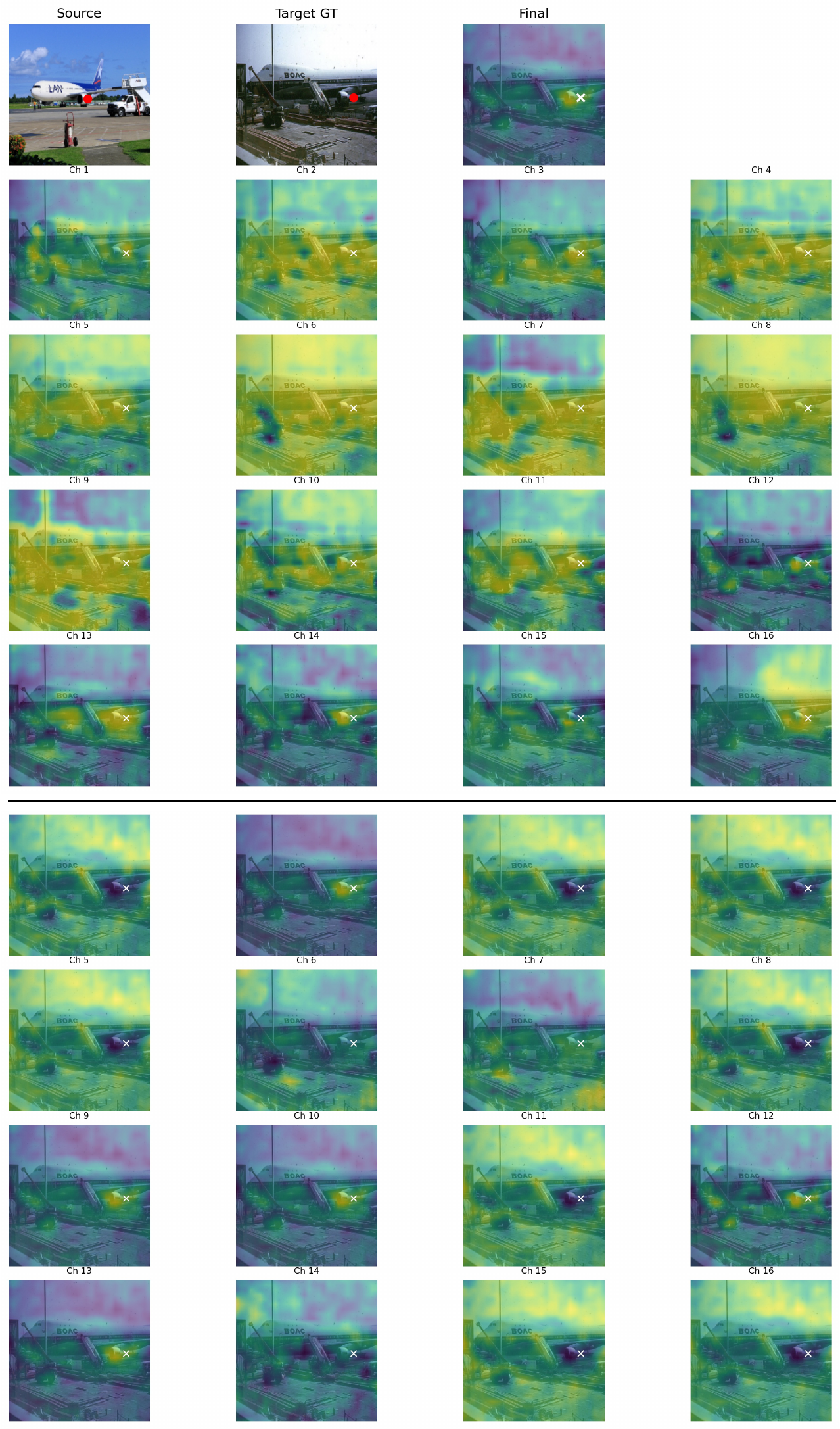}
    \vspace{-3mm}
    \caption{\textbf{Additional visualization of similarity-aware selective scan of \methodName}.
(First row) Source and target images with GT keypoints shown in red, and our final correlation tensor.
(First block of 4x4) The pre-refinement correlation maps, per channel. (Second block of 4x4) The post-refinement correlation maps, per channel.
As observed, after refinement, each correlation map is refined to be either \textit{perfectly} wrong, \ie, high attention everywhere other than the GT position, or accurately reflecting the keypoint position.
This visualization helps illustrate that during the final prediction of $\hat{\mathbf{C}}$ using linear projection, the wrong maps are effectively disregarded, and the accurate maps are primarily weighted for aggregation, resulting in our final accurate correlation map.}
    \label{fig:appendix_mamba_qual_2}
    \vspace{-3mm}
\end{figure*}

We provide additional visualizations of refined correlations in~\cref{fig:appendix_mamba_qual_1} and ~\cref{fig:appendix_mamba_qual_2}.
Fig.~\ref{fig:mamba_viz} demonstrates that our refined correlation map can better localize keypoints -~\cref{fig:appendix_mamba_qual_1} and ~\cref{fig:appendix_mamba_qual_2} aim to provide a deeper insight into this phenomenon.
In~\cref{fig:appendix_mamba_qual_1} and ~\cref{fig:appendix_mamba_qual_2}, the top-left images represent an image pair with a ground truth correspondence. The top-right image visualizes the output correlation map from \methodName. 
This visualization helps illustrate that during the final prediction of $\hat{\mathbf{C}}$ using linear projection, the wrong maps are effectively disregarded, and the accurate maps are primarily weighted for aggregation, resulting in our final accurate correlation map.

\section{Correlation robustness analysis}
\label{sec:appendix_correlation_robustness}

\begin{figure*}[t]
    \centering
    \includegraphics[width=1.0\textwidth]{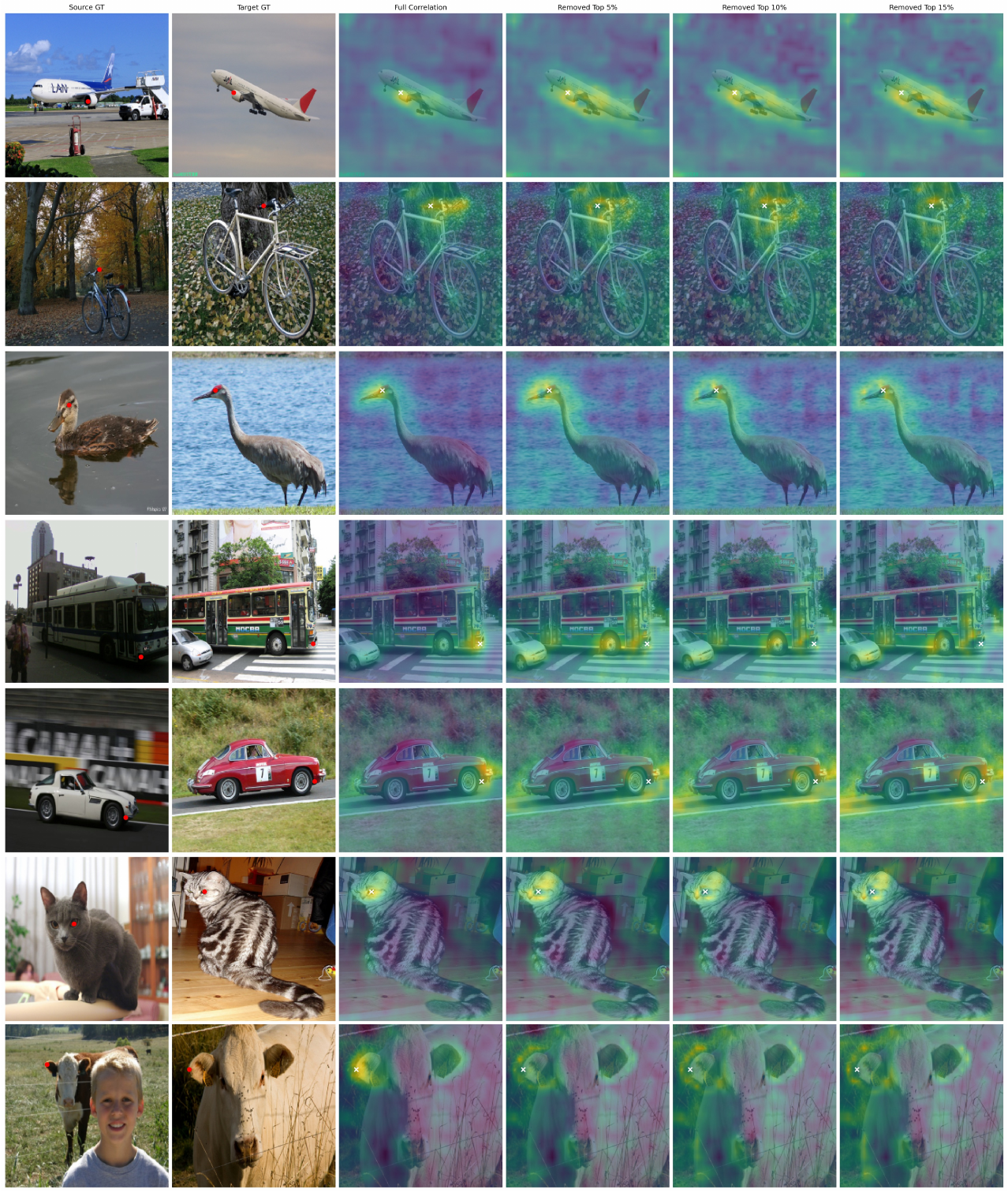}
    \vspace{-3mm}
    \caption{\textbf{Correlation robustness analysis}.
    The first two columns show the source and target images with a GT keypoint pair, followed by the results of our refinement after removing 0\% (ours), 5\%, 10\%, and 15\% of the entries with the highest correlation scores.
    It can be seen that the localization of target keypoint degrades when the top-$k\%$ of the correlation scores are replaced with zeros, showing progressively worse localization with higher $k$.
    This evidences that establishing a strong and reliable contextual foundation based on accurate matches (\ie, highest correlation score) yields strong benefits to the correlation refinement process.
}
    \label{fig:top_k_ablation}
\end{figure*}

We aim to provide insights into how the model's correlation refinement process depends on high-confidence correspondences, by replacing the top-$k\%$ of the correlation scores with zeros in~\cref{fig:top_k_ablation}. 
It shows that the higher the $k$, \ie, more high-confidence correlation values are removed, the localization performance degrades more. 
This evidences that the correlation refinement process relies heavily on the strongest initial correspondences to produce accurate final predictions.

\section{Cumulative contribution analysis of each correlation state}
\label{sec:cumulative_contribution}

\begin{figure*}[t]
    \centering
    \includegraphics[width=0.75\textwidth]{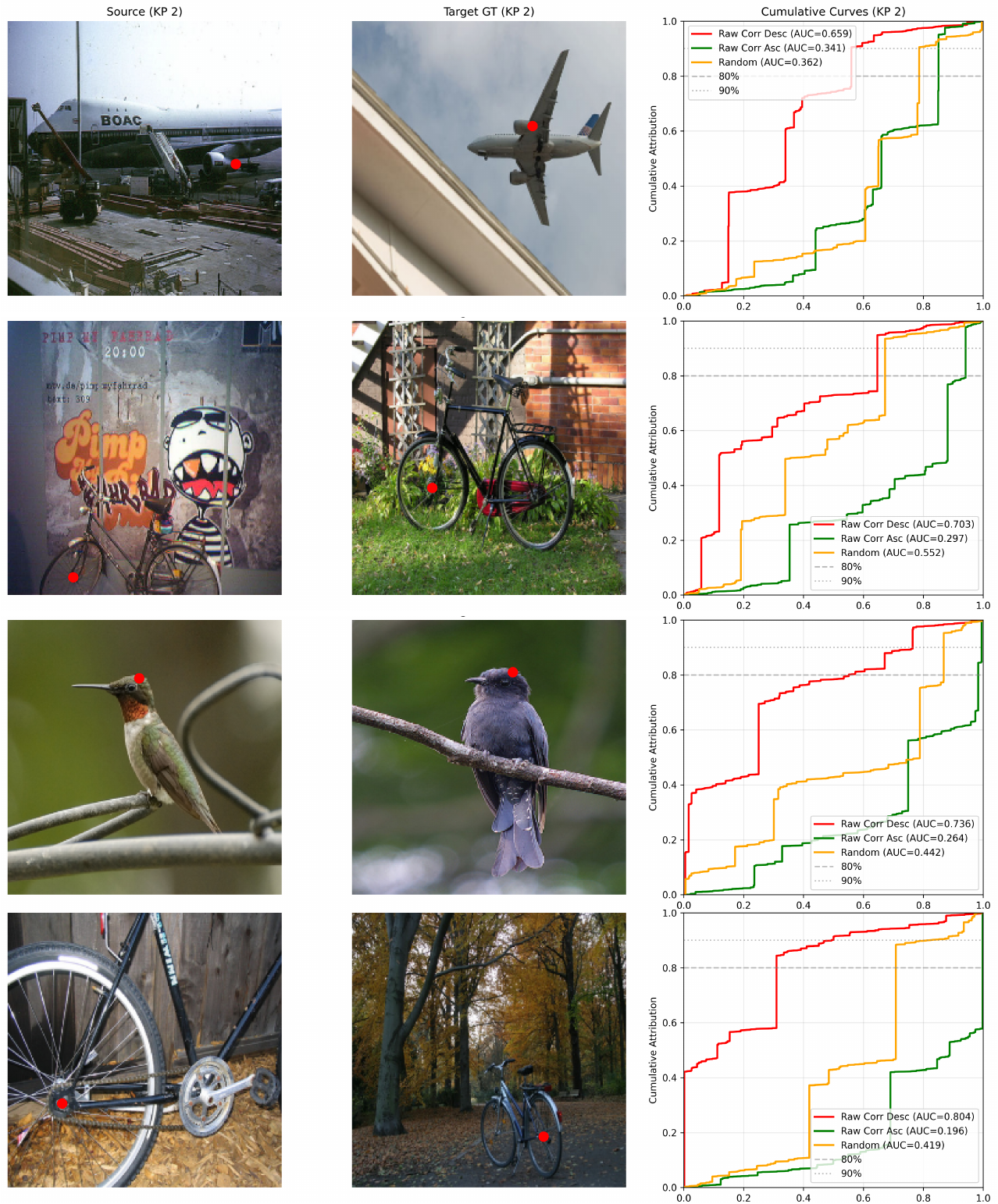}
    \vspace{-3mm}
    \caption{\textbf{Cumulative attribution analysis}.
    The visualization presents three panels for each keypoint: Source - Target - Cumulative Curves. 
    We use integrated gradients to compute attribution weights for each correlation token, in order to reveal how quickly the model accumulates useful evidence when tokens are ordered by different criteria.
    The cumulative curves show the proportion of total attribution mass accumulated as a function of token fraction, with Area Under Curve (AUC) metrics quantifying concentration efficiency. 
    Higher AUC values indicate that important evidence is concentrated in fewer tokens, while lower values suggest more distributed attribution.
    This visualization illustrates that traversing the tokens in the descending order of correlation scores achieves higher concentration of useful evidence compared to ascending or random orderings, demonstrating that the model's refinement process effectively leverages high-confidence initial correspondences.
}
    \label{fig:cumulative_contribution}
\end{figure*}

In~\cref{fig:cumulative_contribution}, we visualize the cumulative contribution analysis of each correlation state, \ie, each entry of the correlation sequence, depending on how we order the correlation sequence.
We use integrated gradients to compute attribution weights for each correlation token, in order to reveal how quickly the model accumulates useful evidence when tokens are ordered by different criteria.
The cumulative curves show the proportion of total attribution mass accumulated as a function of token fraction, with Area Under Curve (AUC) metrics quantifying concentration efficiency.   
Higher AUC values indicate that important evidence is concentrated in fewer tokens, while lower values suggest more distributed attribution.
This visualization illustrates that traversing the tokens in the descending order of correlation scores achieves higher concentration of useful evidence compared to ascending or random orderings, demonstrating that our refinement process effectively leverages high-confidence initial correspondences.

\section{Additional qualitative results}
\label{sec:appendix_qualitative_results}

\begin{figure*}[t]
    \centering
    \includegraphics[width=1.0\textwidth]{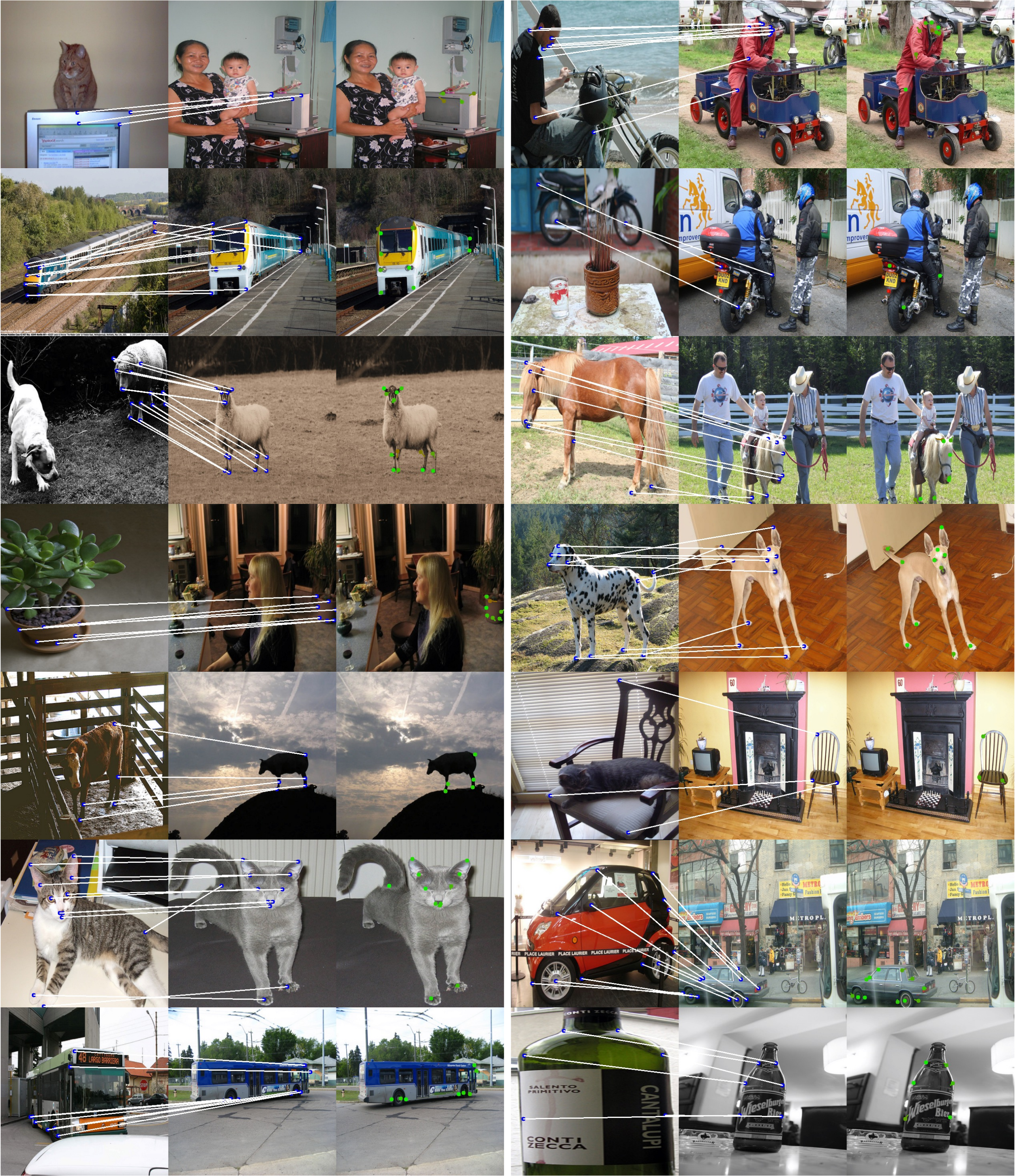}
    \caption{\textbf{Additional qualitative results of \methodName}. Best viewed on electronics, when zoomed-in. 
    The left two columns visualize the ground-truth correspondences. The third column visualizes the predicted target keypoints, and its deviation from the GT keypoints.
} 
    \label{fig:appendix_qual}
    \vspace{-3mm}
\end{figure*}

We provide additional qualitative results in Fig.~\ref{fig:appendix_qual}.

\end{document}